\documentclass[10pt,twocolumn,letterpaper]{article}

\usepackage{cvpr}

\usepackage[dvipsnames]{xcolor}
\usepackage{colortbl}
\definecolor{tabbest}{RGB}{255,190,190}
\definecolor{tabsecond}{RGB}{255,220,170}
\newcommand{\best}[1]{\cellcolor{tabbest}#1}
\newcommand{\snd}[1]{\cellcolor{tabsecond}#1}
\usepackage{algorithm}
\usepackage{algorithmic}
\usepackage{multirow}
\usepackage{comment}
\usepackage{tikz}
\usetikzlibrary{calc,spy}
\usetikzlibrary{arrows.meta,positioning,calc,shapes.geometric,decorations.pathreplacing,decorations.pathmorphing}

\definecolor{crimson}{rgb}{0.86, 0.08, 0.24}
\definecolor{gray}{rgb}{0.5,0.5,0.5}
\definecolor{green}{rgb}{0, 0.4, 0}
\definecolor{orange}{rgb}{1, 0.5, 0}
\definecolor{mahogany}{rgb}{0.75, 0.25, 0.0}
\definecolor{purple}{rgb}{0.6, 0, 0.6}
\definecolor{darkgreen}{rgb}{0, 0.4, 0}
\definecolor{frenchblue}{rgb}{0.0, 0.45, 0.73}
\definecolor{blue}{rgb}{0.0, 0.0, 0.65}
\definecolor{red}{rgb}{1,0,0}
\definecolor{yellow}{rgb}{1,1,0}
\definecolor{magenta}{rgb}{1,0,1}
\definecolor{pink}{rgb}{1,0.412,0.706}
\definecolor{cyan}{rgb}{0.25,0.55,0.70}
\definecolor{newgreen}{rgb}{0, 0.6, 0.2}

\newlength\paramargin
\newlength\figmargin
\newlength\subfigmargin
\newlength\subsecmargin
\newlength\tabmargin
\newlength\eqmargin

\newlength\presecmargin
\newlength\secmargin

\newlength\rulelength
\definecolor{cvprblue}{rgb}{0.21,0.49,0.74}
\usepackage[pagebackref,breaklinks,colorlinks,citecolor=cvprblue]{hyperref}

\def\paperID{517}
\def\confName{3DV\xspace}
\def\confYear{2027\xspace}

\def \submission {}

\ifx \submission \undefined

    \newcommand{\chen}[1]{\textcolor{red}{[Chen]: #1}}
    \newcommand{\deva}[1]{\textcolor{red}{[Deva]: #1}}
    \newcommand{\changil}[1]{\textcolor{red}{[Changil]: #1}}

\else

    \newcommand{\chen}[1]{{}}
    \newcommand{\deva}[1]{{}}
    \newcommand{\changil}[1]{{}}

\fi

\newcommand{\LineComment}[1]{\STATE \textcolor{orange}{\footnotesize // #1}}

\title{Eulerian Motion Reconstruction for Water Scenery}

\author{
Chuhan Chen$^{1,*}$ \quad
Yen-Chi Cheng$^2$ \quad
Ayush Saraf$^3$ \quad
Rajvi Shah$^3$ \quad
Tuotuo Li$^3$ \quad
Johannes Kopf$^3$ \quad
\\[2pt]
Chen Gao$^{3,\dagger}$ \quad
Hung-Yu Tseng$^{3,\dagger}$ \quad
Deva Ramanan$^1$ \quad
Matthew O'Toole$^1$ \quad
Changil Kim$^3$
\\[10pt]
$^1$Carnegie Mellon University \quad
$^2$University of Illinois Urbana-Champaign \quad
$^3$Meta 
\\[2pt]
{\tt\small \href{https://sally-chen.github.io/eulersplats/}{sally-chen.github.io/eulersplats/}}
}

\begin{document}

\twocolumn[{
  \renewcommand\twocolumn[1][]{#1}
  \maketitle
  \begin{center}
    \centering
    \includegraphics[width=0.9\textwidth]{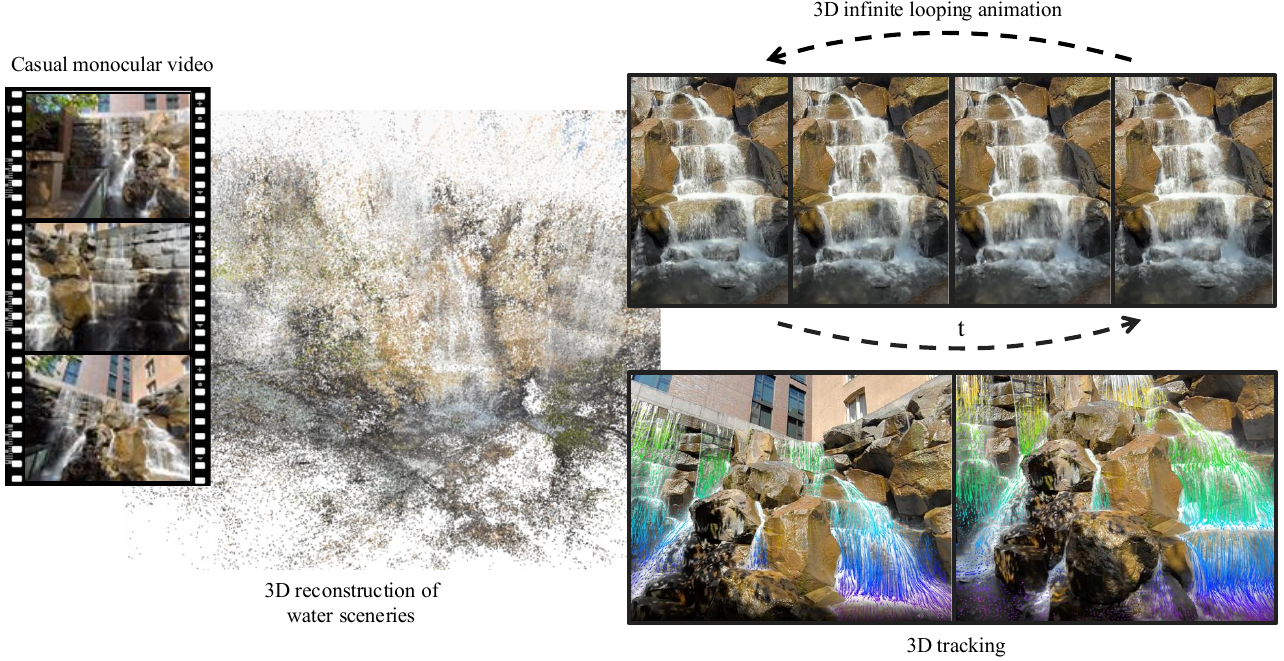}
    \captionof{figure}{Given a casual monocular video of a watery scene, we propose a pipeline to reconstruct a cyclic 4D representation that allows for novel view synthesis, infinitely looping animations, and consistent 3D tracking of water particles.}
    \label{fig:teaser}
  \end{center}
}]
\let\thefootnote\relax\footnotetext{$^*$Work was done while intern at Meta.}
\addtocounter{footnote}{-1}
\let\thefootnote\relax\footnotetext{$^\dagger$Now at Waymo.}
\addtocounter{footnote}{-1}

\begin{abstract}
  Reconstructing and animating water scenery from nature
  produces compelling and immersive visual experiences.
  Previous work examined this task from the perspective of 2D video textures,
  with the goal of creating a looping video.
  In our work, we tackle the problem from a 3D perspective,
  creating a looping 4D dynamic reconstruction
  which can be interactively rendered from novel viewpoints
  from a single non-looping 2D source video.
  We represent motion as a 3D static \textit{Eulerian} motion field
  that advects canonical Gaussian splats that are cyclically reborn
  at fixed time periods, supervised using rendering losses.
  To model non-periodic and stochastic dynamics present in real-world scenes,
  we add a non-periodic, time-varying residual term to capture deviations from the static Eulerian motion field.
    We show quantitatively and qualitatively that our framework enables photorealistic animation of water scenes
  better than prior art. 
  \vspace{-10pt}
\end{abstract}

\section{Introduction}

\label{sec:intro}
Water scenery is ubiquitous in nature, 
from lakes to flowing rivers and waterfalls.
 Capturing and reconstructing such dynamics in a photorealistic manner 
allows virtual 3D worlds to come to life,
  creating immersive experiences. 
  To enable the scalable creation of such experiences, we address the problem of creating 4D reconstructions from casual monocular 2D video captures of water scenes. Modeling this type of motion in casual videos is challenging due to the constant introduction of new (water) particles and the complex interplay between repetition and stochasticity. 

While rigid and non-rigid object motion has been extensively studied 
in 4D reconstruction \cite{som2024, lei2024mosca, Wu_2024_CVPR, pumarola2020d},
 such reconstructions tend to focus on scenes with persistent objects, where dynamic objects tend to remain in the field of view and few objects enter/leave the field of view.
  Here, scene motion can be naturally modelled as spatial deformation~\cite{pumarola2020d} 
  from a canonical 3D reconstruction, often represented as Gaussian splats~\cite{kerbl3Dgaussians} 
  or a neural field~\cite{mildenhall2020nerf, mueller2022instant}.

However, water scenes are naturally modeled with fluid particles that rapidly flow into and out of view. We address this challenge with a {\em periodic} 4D scene model, \ie, we model the 3D scene with a composition of dynamic particles that are looped every $L$ time steps.
   Our approach is inspired by work for creating looping 2D videos
   \cite{schodl2000videotextures, liao2013automated,Holynski_2021_CVPR} but extended to 3D. In particular, previous methods for creating looping 3D representations often make use of multiple video captures from multiple fixed viewpoints~\cite{thonat2017videobased, videoloop}. 
In contrast, we focus on the more challenging (but more scalable) setting of monocular video capture, where a single time instance is observed from only a single view.

Our key contribution lies in a 
novel 4D representation that models water dynamics
 as a combination of periodic and non-periodic residual components.
  We represent motion as a 3D static \textit{Eulerian} motion field that advects
  canonical Gaussian splats, enabling the formation of a looping animation 
  that captures the dominant periodic behavior.
  
While many state-of-the-art methods 
   rely on long-range 2D particle tracks as input~\cite{som2024, lei2024mosca}, we initialize our representation with pairwise 2D optical flow, which tends to be much less error prone when tracking water particles. 
    To create natural, seamless animation, we then advect the Gaussians using the Eulerian motion field through a  novel simulation algorithm inspired from particle systems used in traditional computer graphics, so that each Gaussian repeats at a random, different time.
  Importantly, to account for the intrinsic stochasticity and irregularities 
  of water motion in real-world video captures, we introduce a residual term to capture non-periodic motion and appearance changes.
   Our model is trained end-to-end via a differentiable rendering loss, 
   ensuring photorealistic reconstruction and faithful temporal coherence. Once optimized, our representation allows creation of infinitely-long loopable 3D water animations.

Through both quantitative and qualitative evaluations, we demonstrate that our framework produces realistic and visually compelling animations of water scenes, outperforming existing 4D reconstruction baselines in terms of reconstruction and animation quality.

In summary, we contribute: 
\begin{itemize}
    \item A novel 4D looping representation for reconstructing real-world water scenes that may be non-periodic and stochastic.
\item An efficient model for dynamic scene motion that consists of a {\em static} (non--time-varying) Eulerian motion field, initialized from off-the-shelf 2D optical flow and supervised with an ODE solver using a standard rendering loss.
\item A novel Gaussian simulation technique that advects Gaussians through the Eulerian motion field seamlessly.
\item A time-varying residual appearance term for modeling non-periodic appearance variations in real-world scenes.
\end{itemize}

\section{Related Work}
\label{sec:related_works}
Our work sits at the intersection of research areas: looping 2D video generation for scenes containing water scenery, and 3D reconstruction and novel-view synthesis for dynamic scenes. Below, we briefly review both topics.

\vspace{-2pt}\paragraph{Looping Video Cinemagraph.}
There is a unique line of work that tackles water scenery as a looping video texture. Specifically, from 2D videos, \cite{schodl2000videotextures} identifies visually similar frames and creates loops by transitioning between them; \cite{liao2013automated, liao2015fast} optimize for per-pixel start times and loop periods, aiming to maximize spatiotemporal coherence. More recently, there has been work that moves this problem to 3D where they take multiview asynchronous videos and reconstruct a loopable 3D representation by temporally merging asynchronous videos from nearby views \cite{thonat2017videobased} and building a view-consistent looping Multi-Tile Video~\cite{videoloop} representation based on Multiplane Image~\cite{zhou2018stereo} but using a looping loss to accommodate the non-periodic nature of water. While also trying to reconstruct a 3D looping representation, our work takes in monocular capture. Another line of work builds video loops from single images in 2D, leveraging image based priors like optical flow and monocular depth predictors~\cite{Holynski_2021_CVPR,Jin2025Dynamic}, as well as physical simulation~\cite{fan2022simulating} and user input~\cite{mahapatra2020controllable}. Inspired by Holynski~\etal~\cite{Holynski_2021_CVPR}, we model water motion as an Eulerian motion field but in 3D, and tackle the task of reconstructing motion from input video which is different from animating from a single image.

\vspace{-2pt}\paragraph{Novel View Synthesis for Dynamic Scenes.}

Novel View Synthesis (NVS) aims to render unseen perspectives from limited input viewpoints. For dynamic scenes, this requires a 4D representation that captures both geometry and time-varying motion. They are best captured using  synchronized multi-view recordings, but they are often feasible only in studio environments \cite{Lombardi21}. 
With progress in neural rendering leveraging differentiable representations such as Neural Radiance field~\cite{mildenhall2020nerf} and Gaussian Splatting~\cite{kerbl3Dgaussians}, monocular input has become feasible, though the problem remains severely under-constrained due to simultaneous camera and scene motion. To address this, some approaches incorporate additional sensors like time of flight sensors \cite{Attal2021ToRF, Shandilya2023NeuralFieldsStructuredLighting}, while others rely on learned geometric priors \cite{wang2022_3dmoments, Gao2021DynamicMonocular} or impose hand-crafted motion representation and regularizers \cite{park2021nerfies, Shih2024AmbientGS, park2021hypernerf, tretschk2021nrnerf, som2024, lei2024mosca}, which typically only apply to simple and low frequency dynamics and persistent objects where there is minimum introduction of new objects.
In contrast, water can exhibit high frequency and complex dynamics, where off-the-shelf geometry or motion priors used in previous works can fail (\eg, \cite{yang2024depth_anything_v2,  karaev2023cotracker}). Additionally, because of the continuous flow of water, there is frequent introduction of new particles to the scene and disappearing of old particles. 

\vspace{-2pt}\paragraph{Feed-forward 4D reconstruction.} Feed-forward 4D reconstruction methods predict dynamic 3D/4D representations directly from monocular video, avoiding costly per-scene optimization. Approaches such as L4GM~\cite{ren2024l4gm}, DGS-LRM~\cite{lin2025dgslrm}, 4DGT~\cite{xu20254dgt}, 4D-LRM~\cite{ma20254dlrm}, and MoVieS~\cite{lin2026movies} jointly model geometry, appearance, and motion, often using dynamic or 4D Gaussians and supporting novel-view and novel-time rendering. Their main advantages are fast inference and cross-scene generalization, though their learned motion priors are typically trained on object-centric or general dynamic scenes difficult to generalize to highly stochastic phenomena such as fluids. Compute constraints also prevent them from processing large number of frames for sufficient scene coverage. A complementary direction is generative novel view synthesis, where methods such as TrajectoryCrafter~\cite{yu2025trajectorycrafter}, CogNVS\cite{chen2025cognvs} and Vista4D~\cite{lin2026vista4d} use video diffusion models, guided by geometry and target camera trajectories, to synthesize novel-view videos. These methods can produce visually strong results by hallucinating unseen content and correcting reconstruction errors, but they generally output a requested video in the same time horizon rather than a persistent editable and time-interpolatable 4D scene representation.

\begin{figure*}[t]
    \centering
    \includegraphics[width=\textwidth]{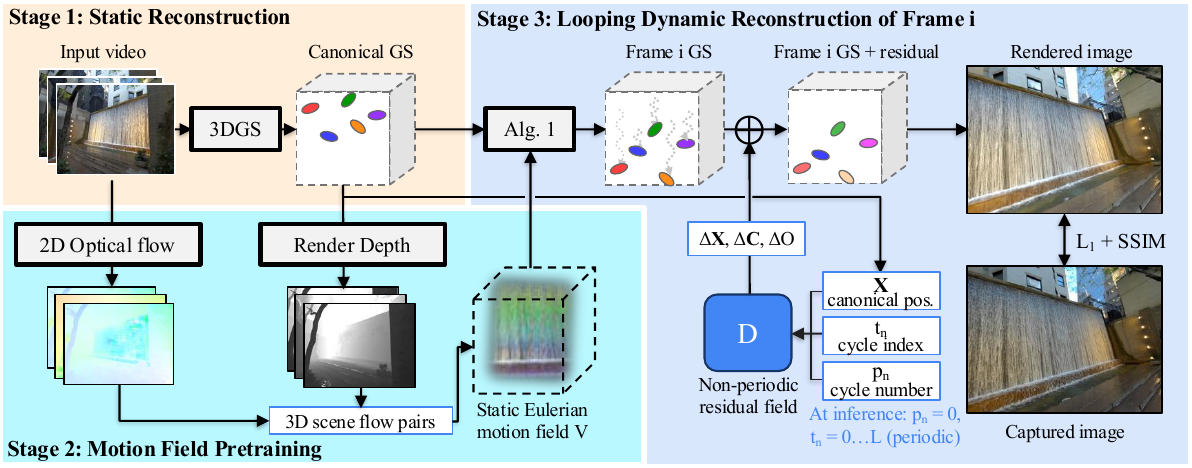}\vspace{-10pt}
    \caption{\textbf{Method Overview.} \textbf{(Stage 1) }Given a monocular input video, we first create a static reconstruction in the form of Canonical Gaussian Splats (Sec.~\ref{subsec:init}). \textbf{(Stage 2)} We use depth computed from the static reconstruction and camera poses to lift predicted 2D optical flow to 3D, and use (3D position, 3D scene flow) pairs at every frame to pre-train a static Eulerian motion field (Sec.~\ref{subsec:init}). \textbf{(Stage 3)} To synthesize input images at each time step, we advect Canonical Gaussian Splats using our motion field through forward Euler integration with our looping algorithm (Sec.~\ref{subsec:looping}, Alg.~\ref{alg1}) and additionally apply a non-periodic residual field $D$ to account for the non-periodic nature of water (Sec.~\ref{subsec:Non-periodic}). Advected splats are rendered into synthesized images; they are compared to the ground truth via rendering losses that are back-propagated to optimize Canonical Gaussian Splat parameters and the motion field (Sec.~\ref{subsec:joint}). Given the trained loopable model, we can generate infinite looping animations by repeating every $L$ frames using Alg.~\ref{alg2} and setting $p_n = 0$.\vspace{-10pt}}
    \label{fig:pipeline}
\end{figure*}

\section{Method}
Our goal is to reconstruct a loopable dynamic representation of a water scene from a monocular video of $T$ frames that captures the scene from different view points and times. The dynamics of water exhibit unique characteristics. The movement of water particles follows a consistent trend when passing through each location in the scene across time, \eg, the flow is fast at the top of a waterfall, and slows in the pool at the bottom. We model the water particles with a canonical set of Gaussian splats~\cite{kerbl3Dgaussians}, and the motion of water particles when passing through each location in the scene with a static Eulerian motion field (Sec.~\ref{subsec:representation}). Through such a motion field, the positions of canonical Gaussians are advected repeatedly over a fixed cycle, and looped using our simulation algorithm  (Sec.~\ref{subsec:looping}). Although water particles follow a steady flow through a specific location in the scene, their motion at different time steps is stochastic. We model such stochasticity with a residual deformation field that models residual offsets in Gaussian position, opacity, and color within and across cycles (Sec.~\ref{subsec:Non-periodic}).

We design a multi-stage approach to learn these components, as shown in Fig.~\ref{fig:pipeline}, where we first initialize canonical Gaussians and the Eulerian motion field separately (Sec.~\ref{subsec:init}), and optimize the three components jointly end-to-end using rendering losses (Sec.~\ref{subsec:joint}). The remainder of this section describes the details of our proposed approach.

\subsection{Scene Representation}\label{subsec:representation}
Our scene is represented by a set of $N$ 3D Gaussian splats in a canonical space: $\{{\bf G}_n\}_{n=1}^{N}$, where each Gaussian ${\bf G}_n = ({\bf X}_n,{\bf S}_n,{\bf R}_n,{\rm O}_n,{\bf C}_n)$ is defined by its 3D position $\mathbf{X}_n = (x,y,z)_n$, scale $\mathbf{S}_n$, orientation $\mathbf{R}_n$, opacity $\mathrm{O}_n$ and color $\mathbf{C}_n$. We represent the motion of water bodies in the scene using an implicit 3D Eulerian motion field  $\mathbf{V}(x,y,z)$ that is fixed for all time.
Intuitively, $\mathbf{V}(x,y,z)$ is the velocity of a water particle when it reaches the point ${\bf X}=(x,y,z)$ in the scene. Importantly, $\mathbf{V}({\bf X})$ is initialized to be the 3D unprojection of optical flow between two adjacent frames.
\\

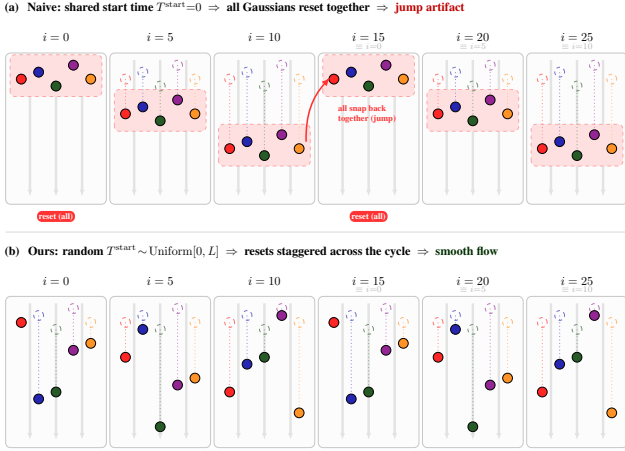
\begin{figure}[t]
\centering
\resizebox{\columnwidth}{!}{
\begin{tikzpicture}[
  >=Latex, font=\small,
  part/.style={circle,draw=black,thick,minimum size=8pt,inner sep=0pt},
]
\newcommand{\panel}[8]{
  \begin{scope}[xshift=#1 cm]
    \draw[rounded corners,gray!45,fill=gray!2] (0.05,-0.1) rectangle (2.95,4.25);
    \foreach \fx in {0.75,1.5,2.25}{
      \draw[gray!22,line width=2pt,-{Latex[length=2mm]}] (\fx,4.05) -- (\fx,0.12);}
    \newcommand\lane[4]{
      \node[circle,dashed,draw=##4!55,minimum size=8pt,inner sep=0pt] at (##1,##2){};
      \draw[##4!45,dotted,thick] (##1,##2) -- (##1,##3);
      \node[part,fill=##4!85] at (##1,##3){};}
    \lane{0.5}{3.5}{#4}{red}\lane{1.0}{3.7}{#5}{blue}\lane{1.5}{3.3}{#6}{green!60!black}
    \lane{2.0}{3.9}{#7}{violet}\lane{2.5}{3.5}{#8}{orange}
    \node[font=\bfseries] at (1.5,4.7) {#2};
    \node[font=\scriptsize\itshape,text=gray!60] at (1.5,4.45) {#3};
  \end{scope}}
\newcommand{\npanel}[8]{
  \begin{scope}[xshift=#1 cm]
    \draw[rounded corners,gray!45,fill=gray!2] (0.05,-0.1) rectangle (2.95,4.25);
    \foreach \fx in {0.75,1.5,2.25}{
      \draw[gray!22,line width=2pt,-{Latex[length=2mm]}] (\fx,4.05) -- (\fx,0.12);}
    \pgfmathsetmacro\ylo{min(#4,#5,#6,#7,#8)}
    \pgfmathsetmacro\yhi{max(#4,#5,#6,#7,#8)}
    \fill[red!12,rounded corners] (0.18,\ylo-0.3) rectangle (2.82,\yhi+0.3);
    \draw[red!35,dashed,thick,rounded corners] (0.18,\ylo-0.3) rectangle (2.82,\yhi+0.3);
    \newcommand\lane[4]{
      \node[circle,dashed,draw=##4!55,minimum size=8pt,inner sep=0pt] at (##1,##2){};
      \draw[##4!45,dotted,thick] (##1,##2) -- (##1,##3);
      \node[part,fill=##4!85] at (##1,##3){};}
    \lane{0.5}{3.5}{#4}{red}\lane{1.0}{3.7}{#5}{blue}\lane{1.5}{3.3}{#6}{green!60!black}
    \lane{2.0}{3.9}{#7}{violet}\lane{2.5}{3.5}{#8}{orange}
    \node[font=\bfseries] at (1.5,4.7) {#2};
    \node[font=\scriptsize\itshape,text=gray!60] at (1.5,4.45) {#3};
  \end{scope}}

\begin{scope}[yshift=7cm]
  \npanel{0}{$i=0$}{}{3.5}{3.7}{3.3}{3.9}{3.5}
  \npanel{3}{$i=5$}{}{2.5}{2.7}{2.3}{2.9}{2.5}
  \npanel{6}{$i=10$}{}{1.5}{1.7}{1.3}{1.9}{1.5}
  \npanel{9}{$i=15$}{$\equiv i{=}0$}{3.5}{3.7}{3.3}{3.9}{3.5}
  \npanel{12}{$i=20$}{$\equiv i{=}5$}{2.5}{2.7}{2.3}{2.9}{2.5}
  \npanel{15}{$i=25$}{$\equiv i{=}10$}{1.5}{1.7}{1.3}{1.9}{1.5}
  \draw[-{Latex[length=2.6mm]},red!80,line width=1.3pt]
    (8.7,1.7) to[bend left=22] (9.4,3.55);
  \node[red!80,font=\scriptsize\bfseries,align=left,anchor=west] at (9.5,2.5)
    {all snap back\\together (jump)};
  \node[fill=red!80,text=white,font=\scriptsize\bfseries,rounded corners,inner sep=1.5pt]
    at (1.5,-0.45) {reset (all)};
  \node[fill=red!80,text=white,font=\scriptsize\bfseries,rounded corners,inner sep=1.5pt]
    at (10.5,-0.45) {reset (all)};
  \node[anchor=west,font=\bfseries] at (-0.1,5.55)
    {(a)\;\, Naive: shared start time $T^{\mathrm{start}}{=}0$ $\;\Rightarrow\;$
     all Gaussians reset together $\;\Rightarrow\;$
     \textcolor{red!75!black}{jump artifact}};
\end{scope}

\begin{scope}[yshift=0cm]
  \panel{0}{$i=0$}{}{3.5}{1.3}{1.5}{2.7}{2.9}
  \panel{3}{$i=5$}{}{2.5}{3.3}{0.5}{1.7}{1.9}
  \panel{6}{$i=10$}{}{1.5}{2.3}{2.5}{3.7}{0.9}
  \panel{9}{$i=15$}{$\equiv i{=}0$}{3.5}{1.3}{1.5}{2.7}{2.9}
  \panel{12}{$i=20$}{$\equiv i{=}5$}{2.5}{3.3}{0.5}{1.7}{1.9}
  \panel{15}{$i=25$}{$\equiv i{=}10$}{1.5}{2.3}{2.5}{3.7}{0.9}
  \node[anchor=west,font=\bfseries] at (-0.1,5.55)
    {(b)\;\, Ours: random $T^{\mathrm{start}}\!\sim\!\mathrm{Uniform}[0,L]$ $\;\Rightarrow\;$
     resets staggered across the cycle $\;\Rightarrow\;$
     \textcolor{green!45!black}{smooth flow}};
\end{scope}

\draw[gray!30] (0.05,6.1) -- (17.95,6.1);
\end{tikzpicture}}
\caption{\textbf{Why per-Gaussian random start times?}
Five Gaussians at different canonical positions (dashed circles) are advected by a
downward Eulerian motion field $\mathbf{V}$ (gray arrows) with cycle length $L{=}15$.
Each is $t_n=(i-t^0_n)\bmod L$ steps below its canonical (rebirth)
position and snaps back on rebirth ($t_n{=}0$).
\textbf{(a)}~With a \emph{shared} start time, every Gaussian has the same $t_n=i\bmod L$
and resets \emph{simultaneously} at $i=0,15,\dots$: the whole scene moves as one rigid
clump (red band) and teleports back to the top in a single frame (jump artifact).
\textbf{(b)}~With \emph{random} start times
($t^0_n=0,3,6,9,12$ for the
\textcolor{red!85}{red}, \textcolor{blue!85}{blue}, \textcolor{green!55!black}{green},
\textcolor{violet}{violet}, \textcolor{orange}{orange} Gaussians) the resets are
staggered across the cycle, so at any frame only a few Gaussians reset while the rest
flow smoothly.}\vspace{-10pt}
\label{fig:looping_comparison}
\end{figure}

\subsection{Looping by Forward Euler Integration }
\label{subsec:looping}

To animate our scenes, we take our canonical Gaussian splats and motion field as input, 
and compute the 3D path that each Gaussian follows over time. Normally, there is an endless supply of particles flowing through the scene, requiring an equal number of Gaussians. Instead, we use a \emph{loopable} representation to model the flow, where the  representation repeats itself after cycle length $L$.   Specifically, Gaussians respawn to the beginning of their path every $L$ steps.  To compute the 3D path each Gaussian follows over 1 cycle, we advect the canonical Gaussian splats forward in time using our Eulerian motion field. 
Specifically, we obtain splat locations at future times $t$ by forward Euler integration using the velocity field $\mathbf{V}$:
\begin{align}
\label{eq:euler}
{\bf X}_{n,t} &= {\bf X}_{n,t-1} +  \mathbf{V}({\bf X}_{n,t-1}) \quad &\forall t \in \{0, \cdots, L\} \text{.}
\end{align}
Note it is possible to continue advecting the Gaussians forward in time indefinitely to model how water truly flows, but this will eventually lead to a scene that is empty of Gaussians since they are advected out of the scene.
Instead,  we respawn (or loop) the Gaussians after every $L$ (``cycle length") frames at their canonical position, which results in an infinitely-long 4D dynamic scene.
The naive way to make the Gaussians loop is to make all the Gaussians re-spawn at the same time, but this produces visual artifacts since relocating all Gaussians at once causes a visible discontinuity in the animation; see Fig.~\ref{fig:looping_comparison}(a)). 
 Instead, we randomly sample the time at which each Gaussian starts moving from a uniform distribution $[0,L]$, 
 which allows each Gaussian to loop by following their distinct schedule and respawn at different times, producing a more natural flow of Gaussians throughout the scene (refer to Fig.~\ref{fig:looping_comparison}(b)).
In this way, we create a ``conveyor belt'' of dynamic Gaussian splats that
  repeat every $L$ time steps.
  We provide a detailed description of our looping algorithm during training time in Alg.~\ref{alg1} and inference time in Alg.~\ref{alg2}.

\begin{figure}[t]
\begin{algorithm}[H]
\caption{
Reconstructing frame~$i$ (training).
\\[1pt]
{\normalfont\footnotesize\emph{Time complexity:}
$O(NL)$ to advance each of $N$ Gaussians up to $L$ Euler steps from its canonical position.}}
\label{alg1}
\begin{algorithmic}[1]
\REQUIRE Canonical Gaussian positions $\{{\bf X}^0_n\}_{n=1}^{N}$,  start times $\{t^0_n\}_{n=1}^{N}$, flow field ${\bf V}$, cycle length $L$\\

    \FOR{$n = 1$ \TO $N$}
     {\LineComment{Compute  cycle index $t$ for Gaussian $n$ at frame $i$.}}
        \STATE $t \gets \left( i - t^0_n \right) \bmod L$ \quad  
         {\LineComment{Take $t$ steps from the canonical position ${\bf X}^0_n$.}}
        \STATE ${\bf {X}}_{n} \gets \mathrm{Propagate}\left({\bf X}^0_n,{\bf V}, t\right)$
    \ENDFOR
    \vspace{2pt}
    \hrule
    \vspace{2pt}
    \STATE \textbf{Subroutine} $\mathrm{Propagate}({\bf X},{\bf V},t)$:
    \FOR{$k = 1$ \TO $t$}
        {\LineComment{one forward-Euler step, cf.\ Eq.~\ref{eq:euler}}}
        \STATE ${\bf X} \gets {\bf X} + {\bf V}({\bf X})$ \quad 
    \ENDFOR
    \STATE \textbf{return} ${\bf X}$

\end{algorithmic}
\end{algorithm}\vspace{-20pt}
\end{figure}

\begin{figure}
    \begin{algorithm}[H]
\caption{Reconstructing all $L$ frames (inference).\\[1pt]
{\normalfont\footnotesize\emph{Time complexity:} $O(NL)$ to initialize the first frame plus $O(N)$ per subsequent frame.}}
\label{alg2}
\begin{algorithmic}[1]
\REQUIRE Canonical Gaussian positions $\{{\bf X}^0_n\}_{n=1}^{N}$, start times $\{t^0_n\}_{n=1}^{N}$, flow field ${\bf V}$, cycle length $L$\\

\FOR{$i = 1$ \TO $L$}
    \FOR{$n = 1$ \TO $N$} 
    \STATE $t \gets \left( i - t^0_n \right) \bmod L$ 
        \IF{$i = 1$}
            {\LineComment{Initialize position for first frame with cycle index $t$}}
             \quad 
            \STATE ${\bf {X}}_{n} \gets \mathrm{Propagate}\left({\bf X}^0_n,{\bf V}, t\right)$ 
        \ENDIF
        \IF{$t = 0$}
            {\LineComment{Reset to canonical position}}
            \STATE ${\bf {X}}_{n} \gets {\bf {X}}^0_{n} $ 
        \ELSE
            {\LineComment{Advance by one step}}
            \STATE ${\bf {X}}_{n} \gets {\bf {X}}_{n} + \mathbf{V}({\bf {X}}_{n})$ 
        \ENDIF
    \ENDFOR
\ENDFOR

\end{algorithmic}
\end{algorithm}\vspace{-20pt}
\end{figure}

 \begin{figure*}[ht]
  \centering
 {\renewcommand{\arraystretch}{0}
\setlength{\tabcolsep}{1pt}
\newcommand{\hdr}[1]{{\fontfamily{ptm}\selectfont\small #1}}
\newcommand{\zcell}[1]{
  \begin{tikzpicture}[inner sep=0pt,outer sep=0pt]
    \node[anchor=south west] (I) {\includegraphics[width=0.163\linewidth]{#1}};
    \path (I.south west) -- (I.south east)
      coordinate[pos=.333] (xL) coordinate[pos=.667] (xR);
    \path (I.south west) -- (I.north west)
      coordinate[pos=.333] (yB) coordinate[pos=.667] (yT);
    \draw[yellow,line width=1.2pt] (xL |- yB) rectangle (xR |- yT);
    \node[anchor=south west,draw=yellow,line width=1.2pt,inner sep=0pt]
      at (I.south west)
      {\includegraphics[width=0.098\linewidth,viewport=388 291 776 583,clip]{#1}};
  \end{tikzpicture}
}
\newcommand{\imgRGB}[2][0px 0px 0px 0px]{\includegraphics[width=0.163\linewidth,trim=#1,clip]{#2}}
\newcommand{\zoomRGB}[4]{
        \begin{tikzpicture}[
    		image/.style={inner sep=0pt, outer sep=0pt},
    		collabel/.style={above=9pt, anchor=north, inner ysep=0pt, align=center},
    		rowlabel/.style={left=9pt, rotate=90, anchor=north, inner ysep=0pt, scale=0.8, align=center},
    		subcaption/.style={inner xsep=0.75mm, inner ysep=0.75mm, below right},
    		arrow/.style={-{Latex[length=2.5mm,width=4mm]}, line width=2mm},
    		spy using outlines={rectangle, size=1.25cm, magnification=3, connect spies, ultra thick, every spy on node/.append style={thick}},
    		style1/.style={cyan!90!black,thick},
    		style2/.style={orange!90!black},
    		style3/.style={blue!90!black},
    		style4/.style={green!90!black},
    		style5/.style={white},
    		style6/.style={black},
        ]
        
        \node [image] (#1) {\imgRGB[0px 0px 0px 0px]{#2}};
        \spy[style5] on ($(#1.center)-#3$) in node (crop-#1) [anchor=#4] at (#1.#4);
        
        \end{tikzpicture}
    }
\centering
\setlength{\tabcolsep}{-1pt}
\begin{tabular}{@{}cccccc@{}}
\hdr{Ground Truth} & \hdr{Ours} & \hdr{MoSca} & \hdr{AmbGS} & \hdr{4DGS} & \hdr{MoVieS} \\[2pt]
\zoomRGB{tmp}{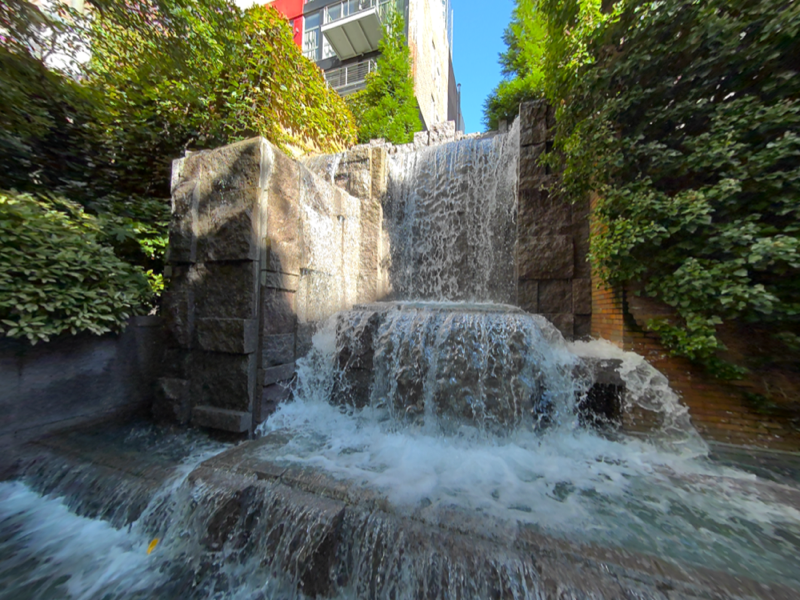}{(0.0,0.1)}{north west} &
\zoomRGB{tmp}{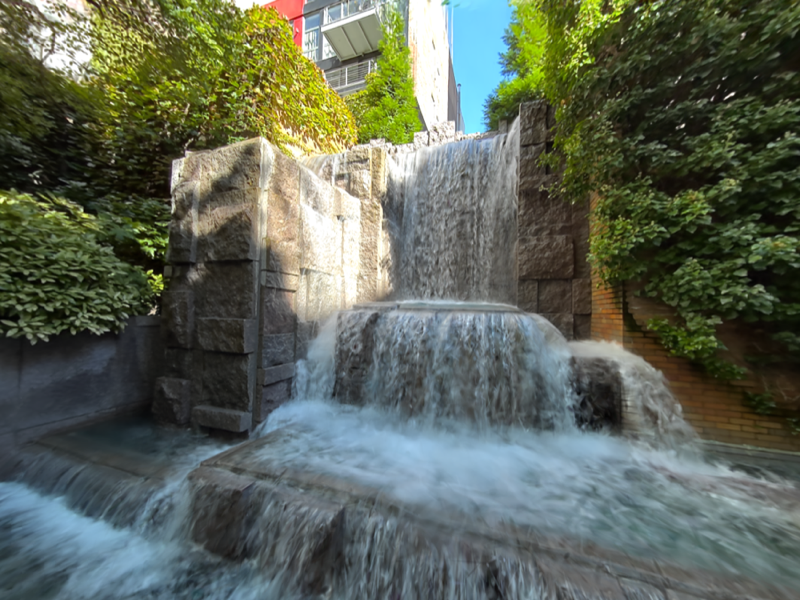}{(0.0,0.1)}{north west} &
\zoomRGB{tmp}{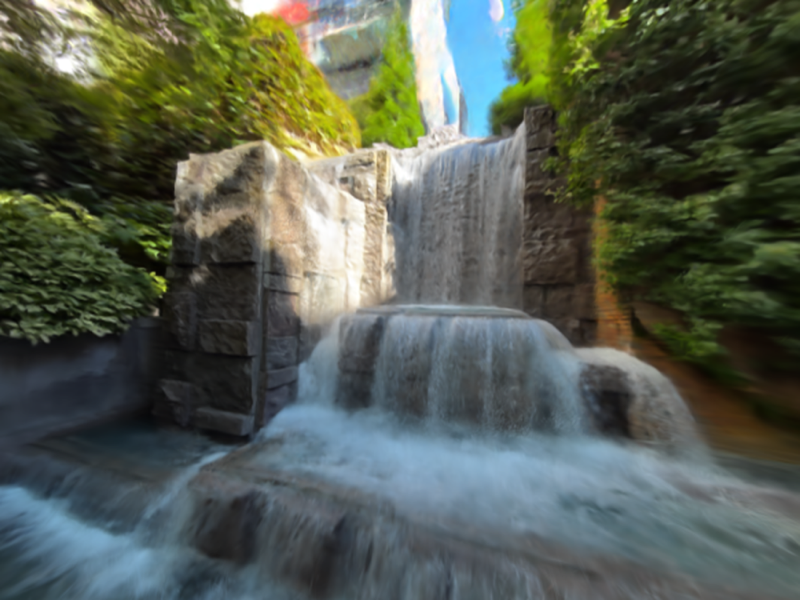}{(0.0,0.1)}{north west} &
\zoomRGB{tmp}{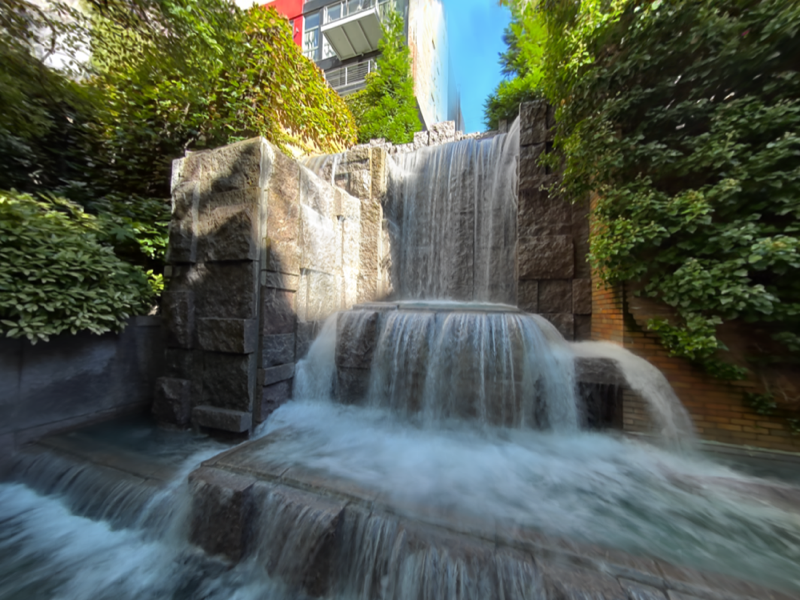}{(0.0,0.1)}{north west} &
\zoomRGB{tmp}{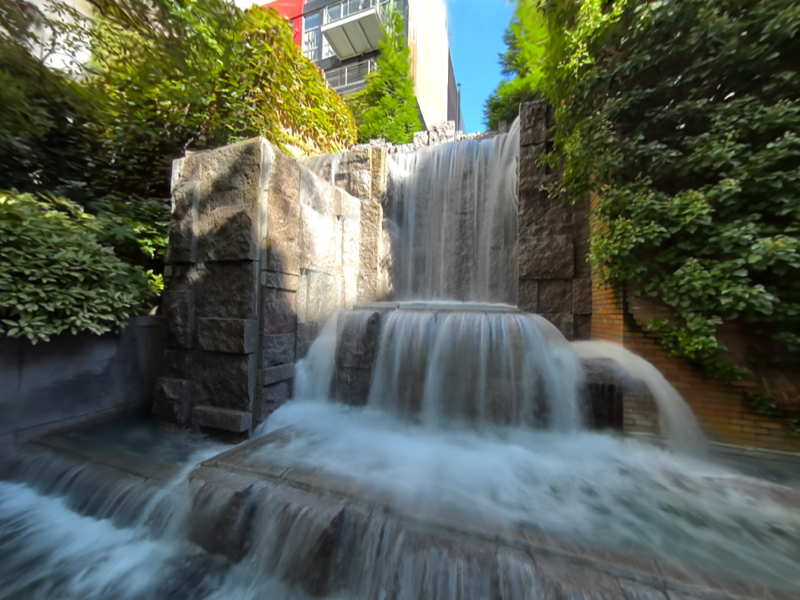}{(0.0,0.1)}{north west} &
\zoomRGB{tmp}{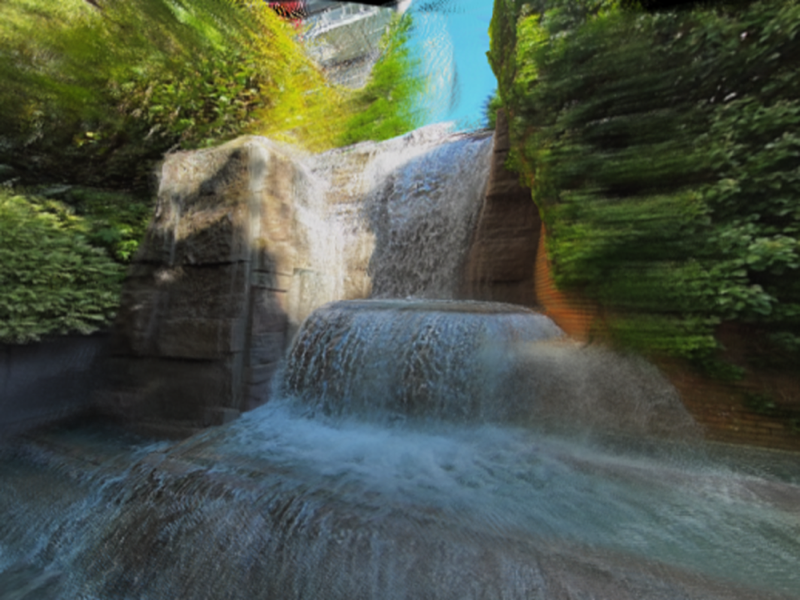}{(0.0,0.1)}{north west} \\

\zoomRGB{tmp}{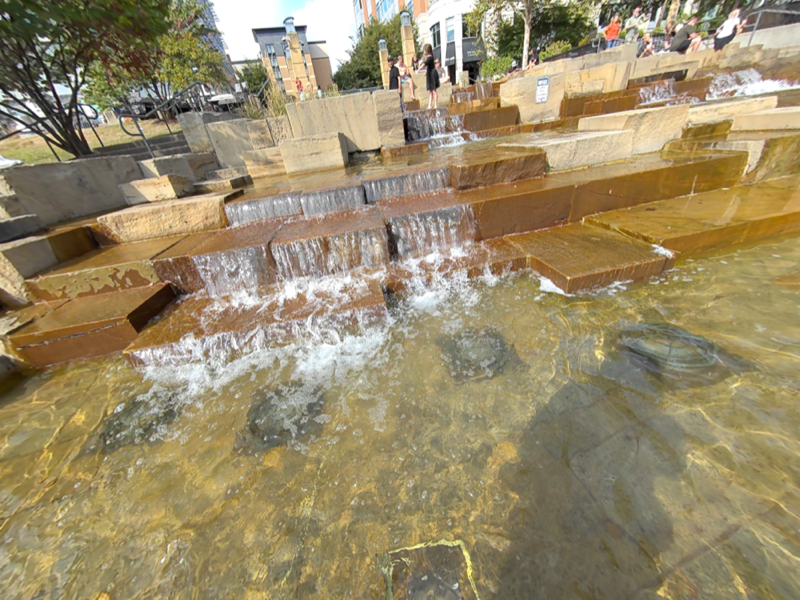}{(0.2,0.2)}{north east} &
\zoomRGB{tmp}{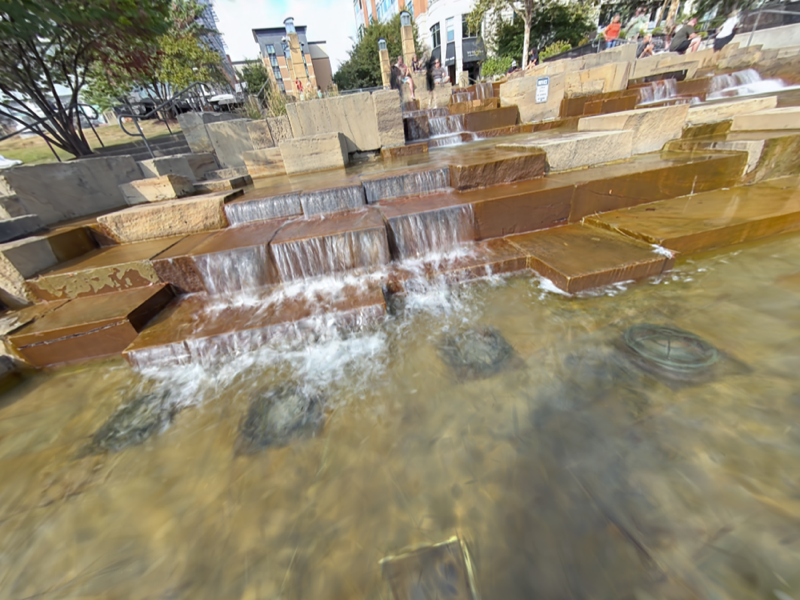}{(0.2,0.2)}{north east} &
\zoomRGB{tmp}{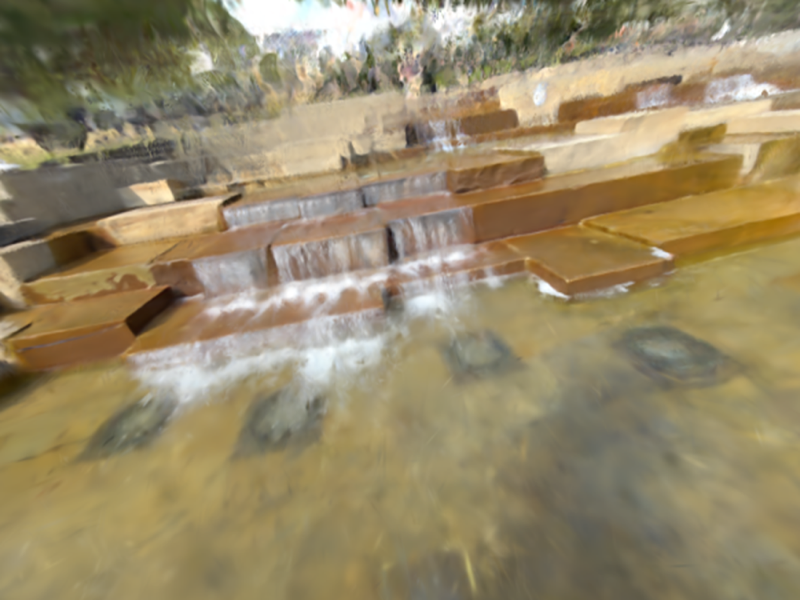}{(0.2,0.2)}{north east} &
\zoomRGB{tmp}{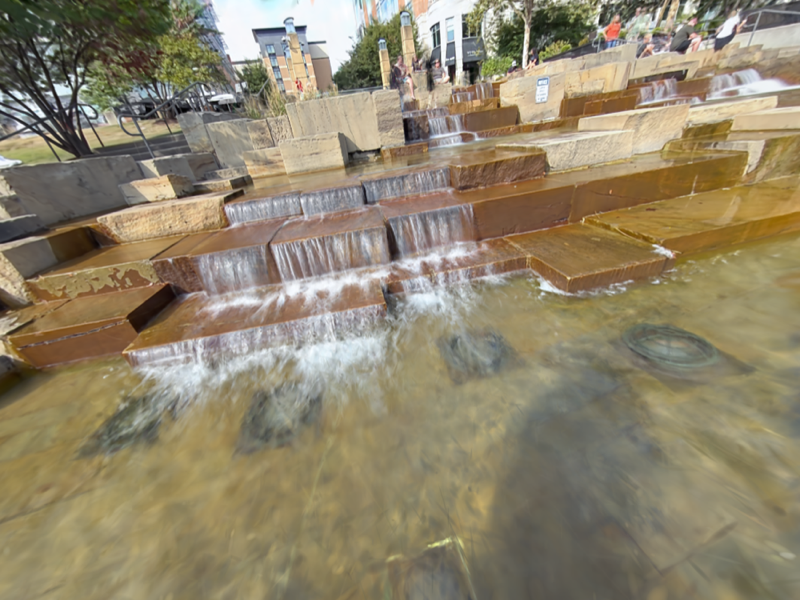}{(0.2,0.2)}{north east} &
\zoomRGB{tmp}{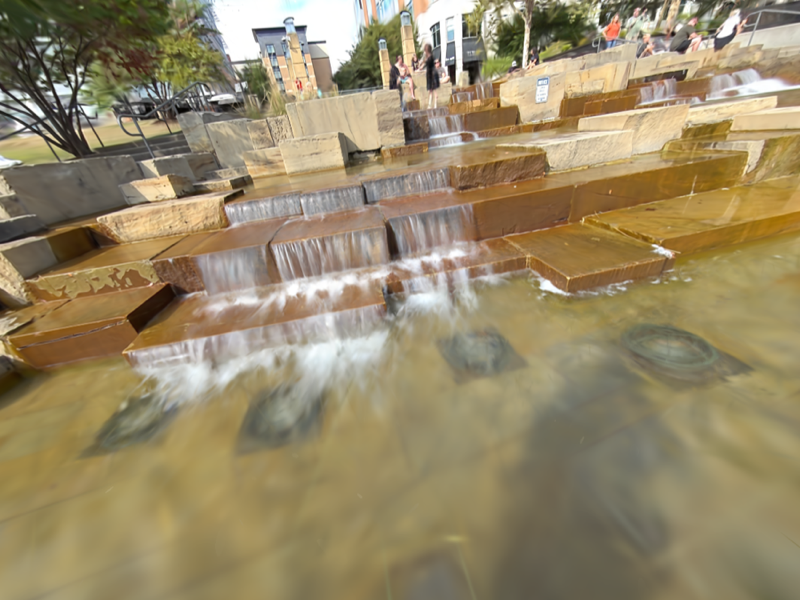}{(0.2,0.2)}{north east} &
\zoomRGB{tmp}{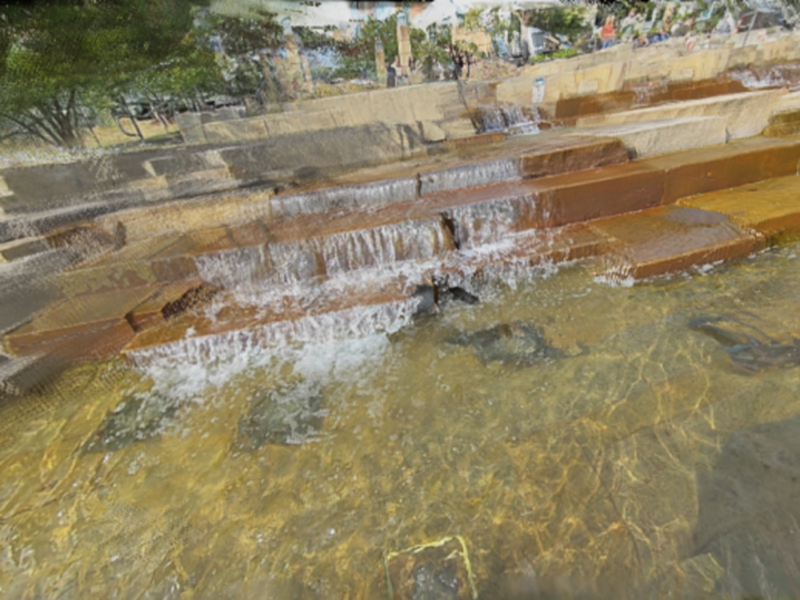}{(0.2,0.2)}{north east} \\

\zoomRGB{tmp}{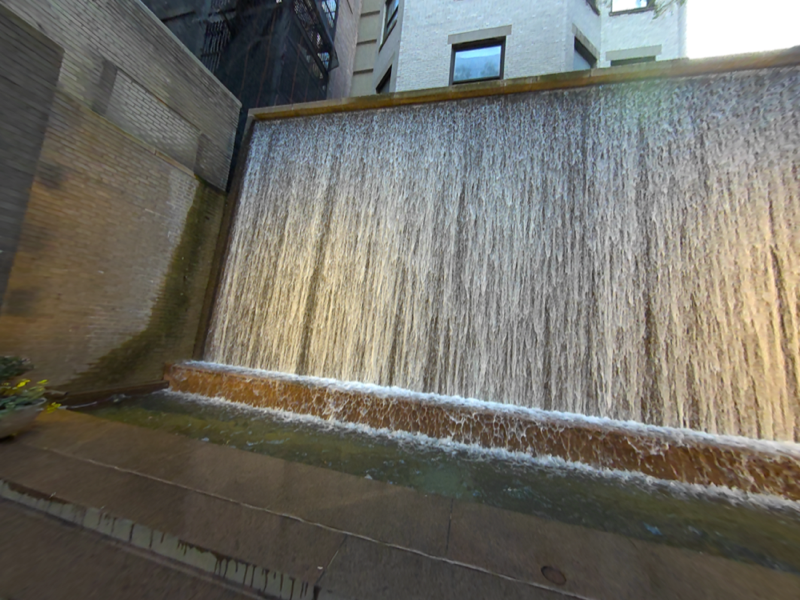}{(-0.4,0.5)}{north west} &
\zoomRGB{tmp}{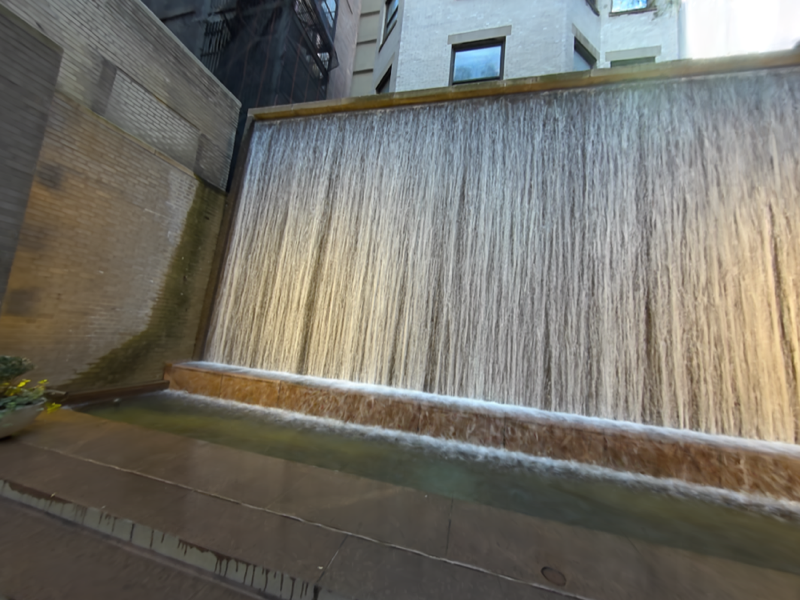}{(-0.4,0.5)}{north west} &
\zoomRGB{tmp}{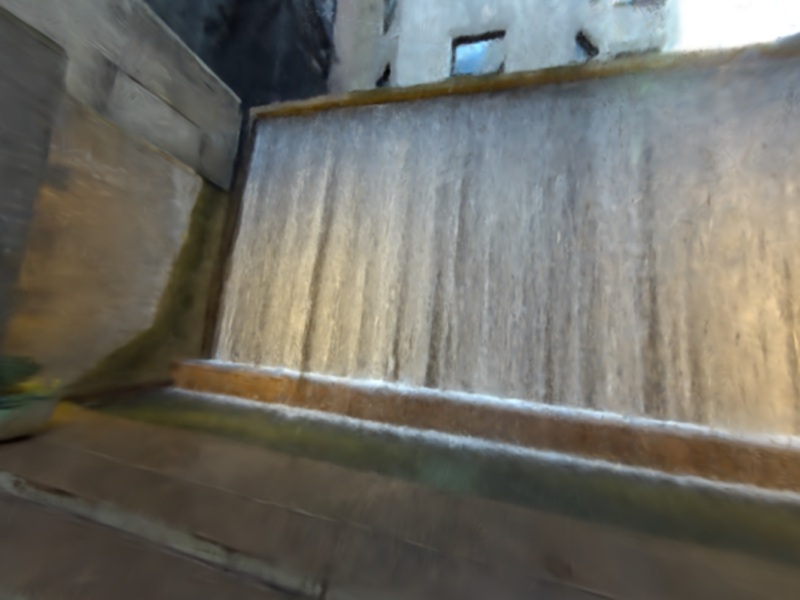}{(-0.4,0.5)}{north west} &
\zoomRGB{tmp}{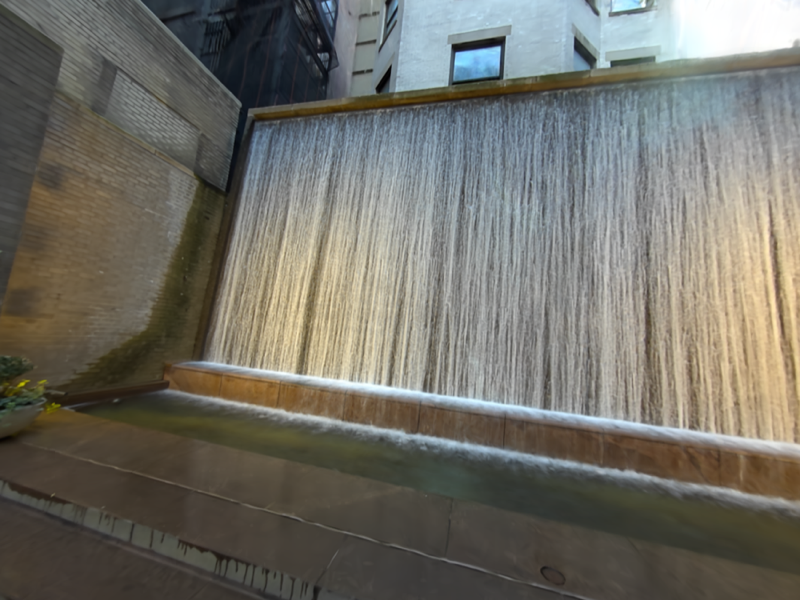}{(-0.4,0.5)}{north west} &
\zoomRGB{tmp}{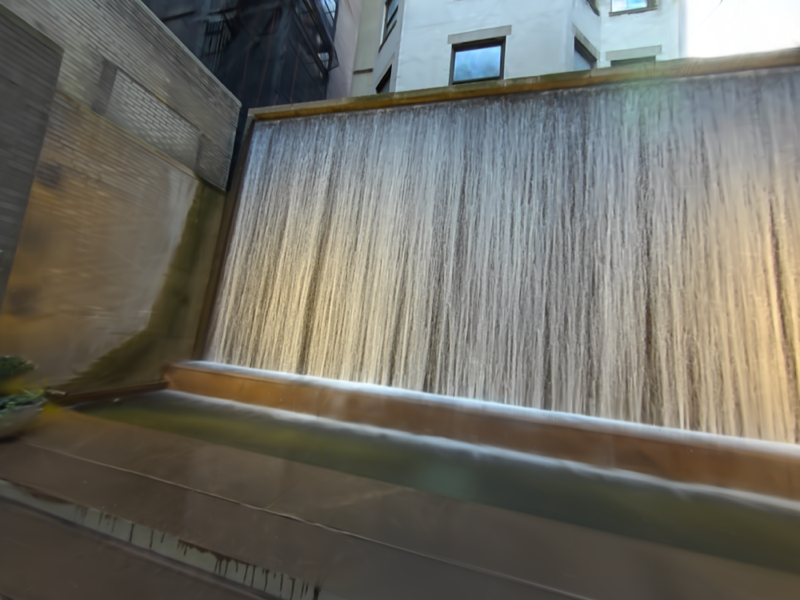}{(-0.4,0.5)}{north west} &
\zoomRGB{tmp}{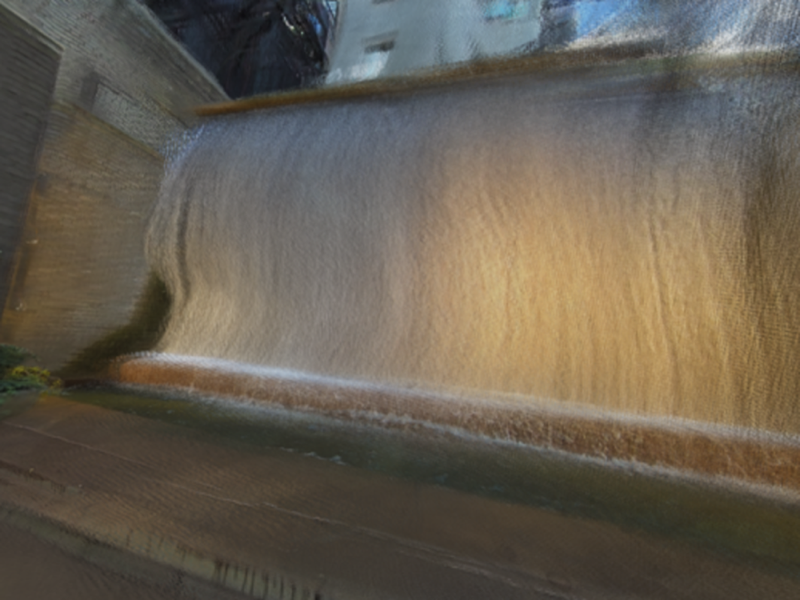}{(-0.4,0.5)}{north west} \\

\zoomRGB{tmp}{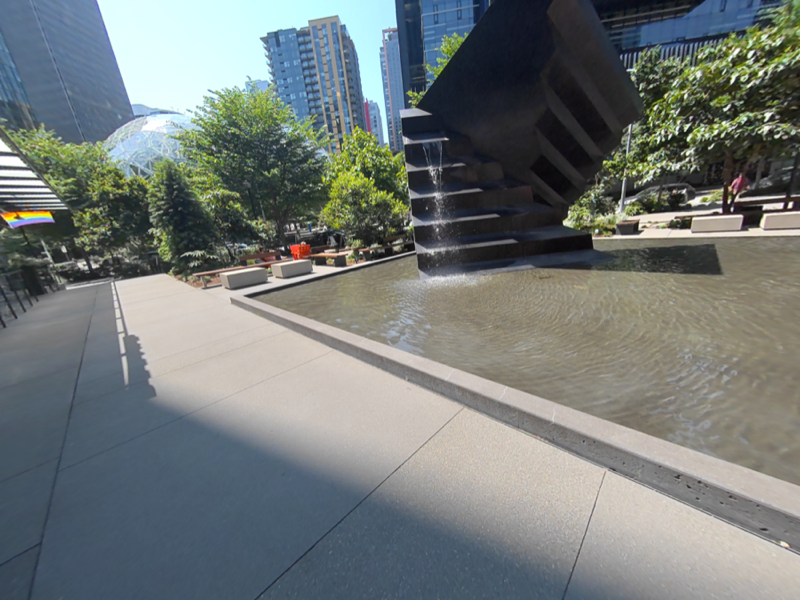}{(-0.6,0.2)}{north west} &
\zoomRGB{tmp}{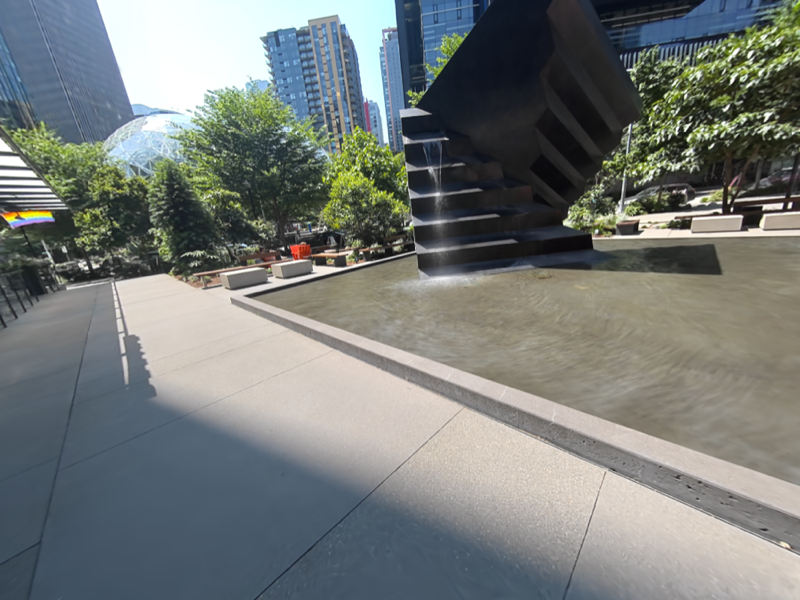}{(-0.6,0.2)}{north west} &
\zoomRGB{tmp}{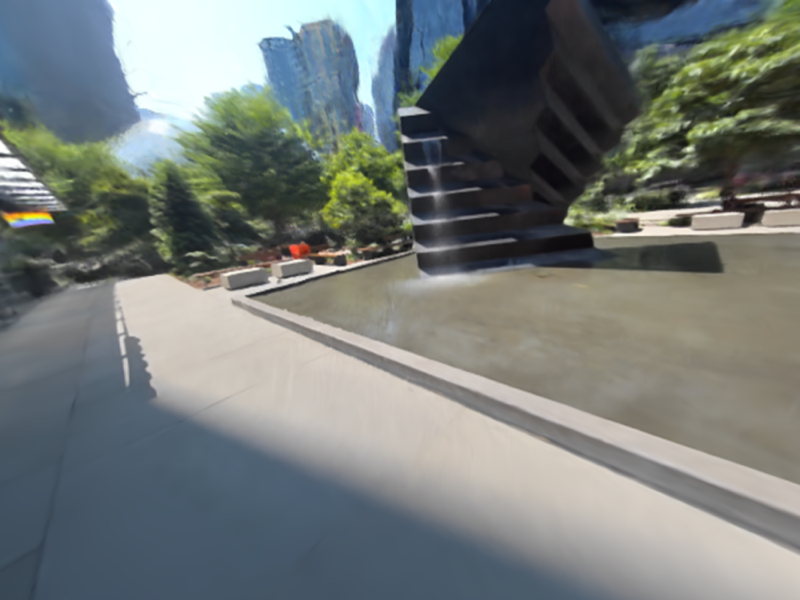}{(-0.6,0.2)}{north west} &
\zoomRGB{tmp}{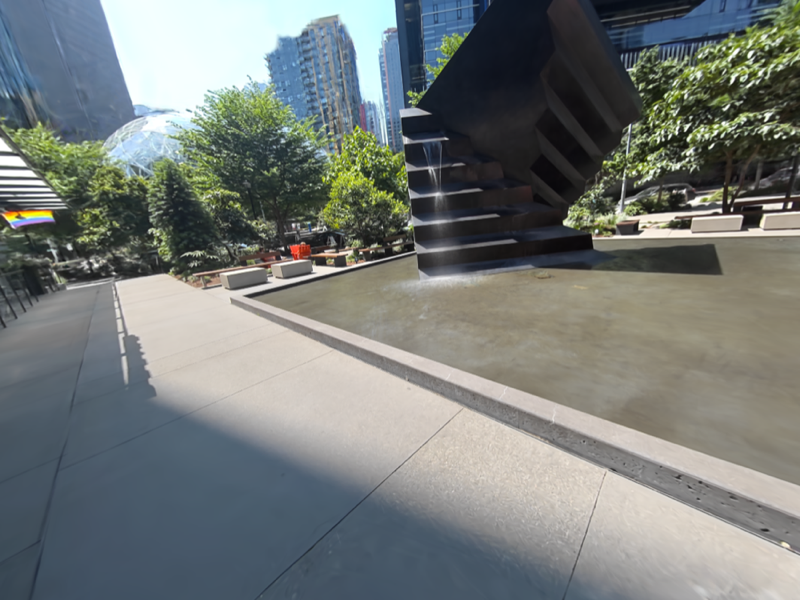}{(-0.6,0.2)}{north west} &
\zoomRGB{tmp}{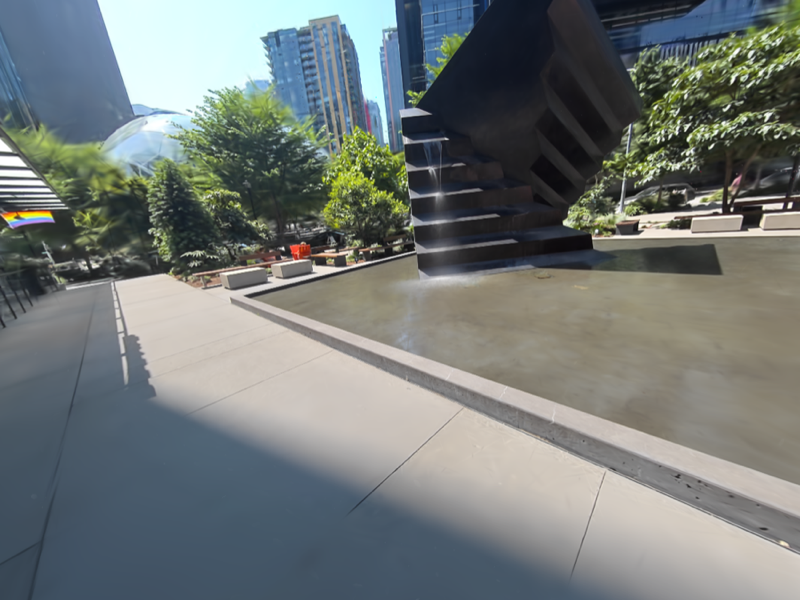}{(-0.6,0.2)}{north west} &
\zoomRGB{tmp}{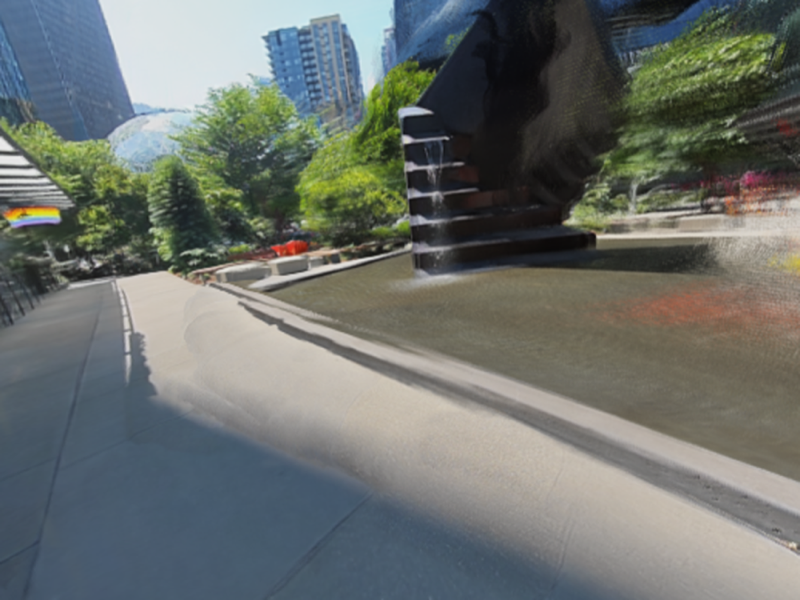}{(-0.6,0.2)}{north west} \\

\end{tabular}}
\vspace{-5pt}
  \caption{\textbf{Qualitative comparison of reconstruction compared to baselines.} Our
method produces much sharper result than the baselines in the water regions. While MovieS appears sharp in row2, its overall view synthesis quality is poor ( refer to videos in the\textbf{ supp. material}).}\vspace{-10pt}
  \label{fig:gt_comp}
\end{figure*}

\subsection{Non-periodic Deformations} \label{subsec:Non-periodic}

The above looping model will generate periodic scenes that exactly repeat every $L$ frames,
 while real-world source videos will naturally contain non-periodic and even stochastic content. 
 This presents a challenge when attempting to optimize a looping model on non-looping data. 
To account for this, we learn a time-varying residual deformation field $\mathbf{D}(\mathbf{X_n}, t_n, p_n)$
that predicts additive offsets to Gaussian position ${\bf X}_n$, Gaussian color ${\bf C}_n$ and opacity ${\rm O}_n$ for the $p_n$th cycle in the input video (cycle number) and  cycle index  $t_n$.
Concretely, for Gaussian \textbf{$G_n$} at frame $i$ with start time $T^{\mathrm{start}}_n$, we factor its elapsed time since birth into a within-cycle \emph{cycle index} $t_n$ and a \emph{cycle number} $p_n$:
\begin{equation}
\label{eq:tp}
t_n = \left( i - T^{\mathrm{start}}_n \right) \bmod L,
\qquad
p_n = \left\lfloor \frac{\,i - T^{\mathrm{start}}_n\,}{L} \right\rfloor .
\end{equation}
The cycle index $t_n$ is the same quantity used in Alg.~\ref{alg1} (the number of forward integration steps since $G_n$'s most recent rebirth), while the cycle number $p_n$ counts how many full cycles $G_n$ were completed in the input video.
These offsets predicted by ${\bf D}$ are {\em not} integrated over time (as velocity-based positional ``offsets'' in Eq.~\ref{eq:euler}),
 but rather are added to a Gaussian's canonical position, color, and opacity to fit the difference between the periodic scene representation with each input frame.
The residual field ${\bf D}$ records the differences in Gaussian parameters across cycles during reconstruction to better fit a training video; its cycle number is set to 0 during inference time so that residuals from the first cycle are used to ensure our 4D animations remain periodic.

\subsection{Initialization}\label{subsec:init}
 Without careful initialization, we find that jointly optimizing Gaussian splats, the Eulerian motion field, and the residual deformation field can easily produce local minima (as shown in Sec.~\ref{subsec:ablations}).
We initialize the canonical splat reconstruction $\{{\bf G}_n\}_{n=1}^{N}$ by fitting a static reconstruction to the input video frames with associated camera poses (inferred via standard SFM toolboxes), ignoring any motion.
To initialize the Eulerian motion field $\mathbf{V}$, we run off-the-shelf optical flow predictor \cite{morimitsu2021ptlflow} on pairs of input frames.
We then lift these 2D optical flow vectors to 3D by using the rendered depth from the static reconstruction. This produces a collection of \emph{pseudo} ground-truth scene flow vectors $\{\bf{V'(X)}\}$ at various locations in the scene, which forms $\{(\bf{X}, \bf{V'(X)})\}$ pairs used to pretrain the Eulerian motion field $\mathbf{V}$. 

\subsection{Joint Training }\label{subsec:joint}
After separately initializing $\{{\bf G}_n\}_{n=1}^{N}$ and $\mathbf{V}$, we jointly train all of $\{{\bf G}_n\}_{n=1}^{N}$, $\mathbf{V}$, and $\mathbf{D}$ using rendering losses. Specifically, we randomly sample a frame $i$ from the input video, propagate the canonical Gaussian splats through $\bf{V}$ to their distinct target cycle index using Alg.~\ref{alg1}. We then query the non-periodic residual field $\bf{D}$ with the cycle index and cycle number corresponding to $i$, and add the residual terms to each Gaussian property. We rasterize the Gaussians using frame $i$'s viewpoint to render the reconstructed frame $i$ and calculate $L_1$ and SSIM losses between the reconstructed and input frame. Additionally, we apply $L_2$ regularizer on the residual field $\bf{D}$ so that it remains small.

\begin{table*}[t]
\centering
\small
\setlength{\tabcolsep}{2.5pt}
\begin{tabular}{lccccccccccccc}
\toprule
& \multicolumn{6}{c}{Full image} & & \multicolumn{6}{c}{Water region} \\
\cmidrule(lr){2-7} \cmidrule(lr){9-14}
Method & PSNR ↑ & SSIM ↑ & LPIPS ↓ & FID ↓ & KID ↓ & FVD ↓ & & PSNR ↑ & SSIM ↑ & LPIPS ↓ & FID ↓ & KID ↓ & FVD ↓ \\
\midrule
MoVieS~\cite{lin2026movies}            & 15.19 & 0.331 & 0.524 & 100.87 & 0.027 & 1564.90 & & 15.35 & 0.291 & 0.526 & 87.21 & 0.014 & 1157.00 \\
MoSca~\cite{lei2024mosca}              & 19.87 & 0.509 & 0.565 & 117.54 & 0.037 &  828.87 & & 23.31 & 0.723 & 0.577 & 104.51 & 0.024 & 534.47 \\
4DGS~\cite{Wu_2024_CVPR}                & \snd{23.01} & 0.693 & 0.397 &  84.00 & 0.026 &  634.19 & & \best{\textbf{25.71}} & \snd{0.803} & 0.436 &  87.58 & 0.022 & 417.40 \\
AmbGS~\cite{Shih2024AmbientGS}   & 22.52 & \snd{0.706} & \snd{0.344} &  \snd{61.32} & \snd{0.014} &  \snd{536.55} & & \snd{25.49} & \best{\textbf{0.806}} & \snd{0.392} &  \snd{69.15} & \snd{0.014} & \snd{314.24} \\
\textbf{Ours}                         & \best{\textbf{23.05}} & \best{\textbf{0.716}} & \best{\textbf{0.315}} & \best{\textbf{39.63}} & \best{\textbf{0.007}} & \best{\textbf{210.01}} & & 25.37 & 0.798 & \best{\textbf{0.375}} & \best{\textbf{46.50}} & \best{\textbf{0.008}} & \best{\textbf{156.69}} \\
\bottomrule\vspace{-16pt}
\end{tabular}
\caption{\textbf{Quantitative comparison of methods.} $\downarrow$ ($\uparrow$) indicates lower (higher) is better. We report metrics over the full image as well as restricted to the water region, which isolates the dynamic content our method targets. FID, KID and FVD follow the distribution-based evaluation protocol of Lift4D~\cite{litman2026lift4d}. Our method achieves the best overall reconstruction accuracy in comparison to the baselines and leads the baselines in spatial and temporal distribution alignment by a large margin.}\vspace{-10pt}
\label{tab:gt_comp}
\end{table*}
\section{Implementation}

Both the Eulerian motion field $\mathbf{V}$ and residual deformation field $\mathbf{D}$  are implemented using triplanes~\cite{zou2024triplane}, where the feature MLP for $\mathbf{V}$ takes only positional features while for $\mathbf{D}$ the feature MLP additionally takes in sinusoid-encoded cycle index and cycle number.
We use an implementation of RayGS~\cite{Rota-Bulo_2025_CVPR} for Gaussian splatting. We first train static reconstruction using canonical Gaussians splats $\{{\bf G}_n\}_{n=1}^{N}$ for 10k iterations. We identify dynamic Gaussians by learning a per-Gaussian dynamic mask from input water segmentation masks.
 We fine-tune the Gaussians along with  $\mathbf{V}$ and $\mathbf{D}$ for another 20k iterations. We choose cycle length $L=15$ for best representation of water dynamics and global appearance (refer to ablation of different L in Sec.~\ref{subsec:ablations})). 
\vspace{-8pt}
\paragraph{Datasets}Due to the lack of monocular dataset dedicated to water scenery, we capture 7 scenes in various environments. Each video records a scene at the resolution of 1168 $\times$ 880 at 30 fps for about 25 seconds with translations and rotations to provide enough parallax. We use COLMAP~\cite{schoenberger2016sfm} to 
reconstruct sparse point cloud and estimate camera parameters. Details on the dataset are included in the \textbf{supp. material}.

\section{Results}
\label{sec:results}

In this section, we present the results of our method with both quantitative and qualitative evaluation. Please see our {\bf supp. material} for rendered results.

\subsection{Quantitative Comparison}

\paragraph{Evaluation data.}
Because of the lack of multiview test data, we follow~\cite{Shih2024AmbientGS} and hold out 3 segments of 30 consecutive
frames per scene from the input video during training, and evaluate exclusively on these \emph{held-out} segments so as to not measure only interpolation between training frames.
\vspace{-13pt}

\paragraph{Baselines.}
We compare our method against four categories that 
(1) relies on dense  view coverage and a deformation field (AmbGS~\cite{Shih2024AmbientGS});
 (2) relies on image-based priors such as 2D tracking and monocular depth (MoSca~\cite{lei2024mosca});
   (3) represents dynamics with generic 4D primitives (4DGS~\cite{Wu_2024_CVPR});
    and (4) reconstructs scene geometry, appearance, and motion with a feed-forward 4D approach (MoVieS~\cite{lin2026movies}). 
  Because MoSca~\cite{lei2024mosca} fails for images bigger than 600$\times$600 resolution in our experiment, we upsample its result to compare with  other methods (see Tab.~\ref{tab:gt_comp} and Fig.~\ref{fig:gt_comp}).
  Additionally, since MoVieS~\cite{lin2026movies} was originally designed for reconstruction from sparse-view short videos and run into memory issues with dense input frames,
  we downsample the input frames to 170 with half resolution to fit into a single A6000 GPU, while providing sufficient coverage and frame rate for the method to reconstruct both geometry and motion of the scene; we also upsample its result to compare with other methods (see Tab.~\ref{tab:gt_comp} and Fig.~\ref{fig:gt_comp}).

\vspace{-15pt}
\paragraph{Full-reference (FR) metrics.}
We report standard per-frame reconstruction quality with PSNR, SSIM and LPIPS.
Since water motion is highly stochastic, it is impossible to fully capture its dynamics with traditional metrics, and metrics like PSNR and SSIM can favor blurry reconstruction \cite{chen2025cognvs}. We
thus also report distribution-based image quality with FID, KID and video quality with FVD following the protocol of \cite{litman2026lift4d, chen2025cognvs,xie2025_sv4d}.
All metrics are reported both on the full image and restricted to the water region, which isolates the dynamic content our method targets.

Aggregated metrics across all scenes, compared against baselines, are presented in Tab.~\ref{tab:gt_comp}. Our method outperforms the baselines on most metrics, especially LPIPS and distributional metrics (FID, KID and FVD) by a big margin showing our method's capability of reconstructing natural and detailed water appearance and motion.
 We further demonstrate this via a user study and qualitative evaluation in the next sections, as well as  video results in our supplementary material.

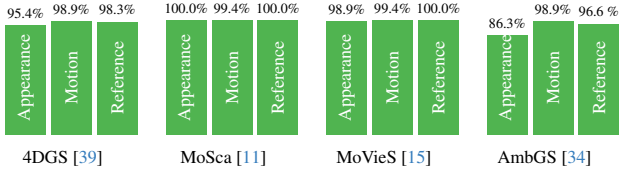
\begin{figure}[t]
  \centering
  \resizebox{\columnwidth}{!}{
  \begin{tikzpicture}[x=1.0cm,y=0.02cm]
  \definecolor{ourgreen}{RGB}{80,180,80}

  \foreach [count=\i from 0] \name/\a/\m/\r in {
    4DGS~\cite{Wu_2024_CVPR}/95.4/98.9/98.3,
    MoSca~\cite{lei2024mosca}/100.0/99.4/100.0,
    MoVieS~\cite{lin2026movies}/98.9/99.4/100.0,
    AmbGS~\cite{Shih2024AmbientGS}/86.3/98.9/96.6
  } {
    \pgfmathsetmacro{\x}{\i*2.8}

    \node[anchor=north,font=\small] at (\x+1.0,-8) {\name};

    \fill[ourgreen] (\x,0) rectangle (\x+0.7,\a);
    \fill[ourgreen] (\x+0.8,0) rectangle (\x+1.5,\m);
    \fill[ourgreen] (\x+1.6,0) rectangle (\x+2.3,\r);

    \node[above,font=\scriptsize] at (\x+0.35,\a) {\a\%};
    \node[above,font=\scriptsize] at (\x+1.15,\m) {\m\%};
    \node[above,font=\scriptsize] at (\x+1.95,\r) {\r\%};
  
  \node[font=\small,white,rotate=90,align=left] at (\x+0.35,50) {Appearance};
  \node[font=\small,white,rotate=90,align=left] at (\x+1.15,50) {Motion};
  \node[font=\small,white,rotate=90,align=left] at (\x+1.95,50) {Reference};
  }\vspace{-10pt}

\end{tikzpicture}}
\vspace{-10pt}
  \caption{\textbf{User study.} Bar height is the percentage of times users rated our
  method's visual quality higher than the competing method (``Winrate''), for three settings:
  \emph{Appearance} (camera moves, scene frozen), \emph{Motion} (camera fixed, scene
  advances), and \emph{Reference} (both advance, shown alongside the real footage).
  Aggregated over $25$ participants and $7$ scenes ($2100$ judgments). Our method was preferred over all baseline methods in all settings.}\vspace{-13pt}
  \label{fig:user_study}
\end{figure}

\subsection{User Study}
\label{subsec:user_study}

We conduct a perceptual user study. We use all $7$ scenes of our dataset and, following~\cite{Shih2024AmbientGS}, for each render two videos per method:
(i) with time fixed at the \emph{held-out} anchor frame while the camera follows the input trajectory, which evaluates reconstructed geometry and appearance;
and (ii) with the camera fixed at the \emph{held-out} anchor viewpoint while time advances, which evaluates reconstructed motion.
We also include a third \emph{reference} setting where both camera and time advance over the \emph{held-out} test frames, shown alongside the ground-truth reference \emph{held-out} footage. Each page presents our result side-by-side with one baseline in all three settings, with left/right order randomized per row and clips served anonymously.
Based on responses from $25$ participants ($2100$ judgments in total),
Fig.~\ref{fig:user_study} reports the rate at which our result was preferred (i.e. “winrate”).
Our method is preferred in $97.6\%$ of all judgments, beating every baseline in every setting. The margin is smallest against AmbGS~\cite{Shih2024AmbientGS} on appearance ($86.3\%$), consistent with it being the strongest prior method on our data; in the motion setting the gap widens to $98.9\%$, indicating that our advantage lies primarily in the recovered dynamics rather than in static appearance.

\relax

\begin{figure}[t]
  \centering
  {\renewcommand{\arraystretch}{0}
   \setlength{\tabcolsep}{0.5pt}
   \newcommand{\hdr}[1]{{\fontfamily{ptm}\selectfont\scriptsize #1}}
   \newcommand{\tcell}[1]{\includegraphics[width=0.195\columnwidth]{#1}}
   \newcommand{\tna}{
     \begin{tikzpicture}
       \node[draw=gray!40,fill=gray!10,rounded corners=1pt,
             minimum width=0.195\columnwidth,minimum height=0.146\columnwidth,
             align=center,inner sep=0pt,text=gray!55,font=\tiny\itshape]{N/A};
     \end{tikzpicture}}
   \begin{tabular}{@{}ccccc@{}}
     \hdr{\textbf{Ours}} & \hdr{4DGS} & \hdr{AmbGS} & \hdr{MoSca} & \hdr{MoVieS} \\[1pt]
     \tcell{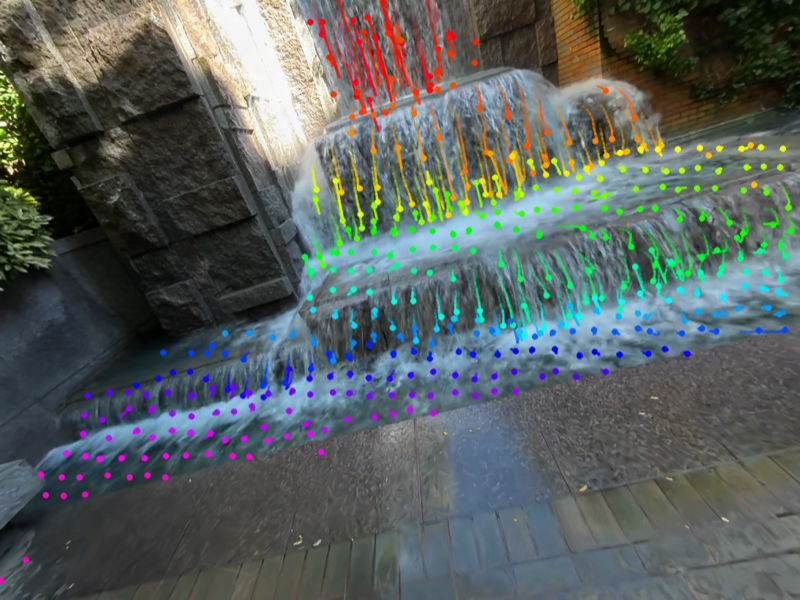}   & \tcell{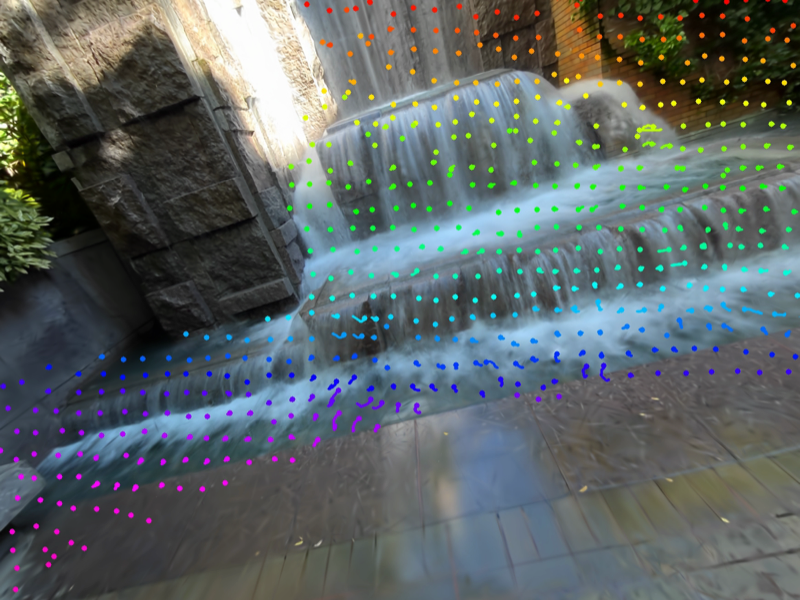}   & \tcell{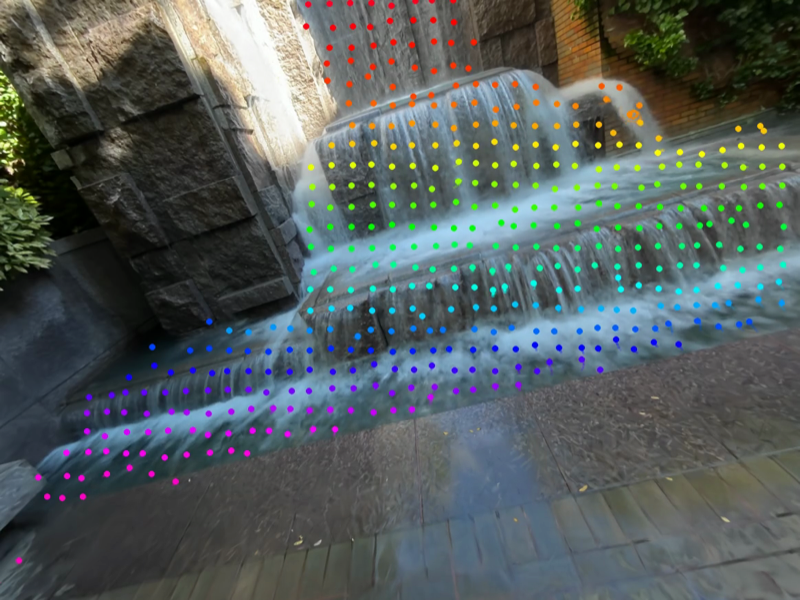}   & \tcell{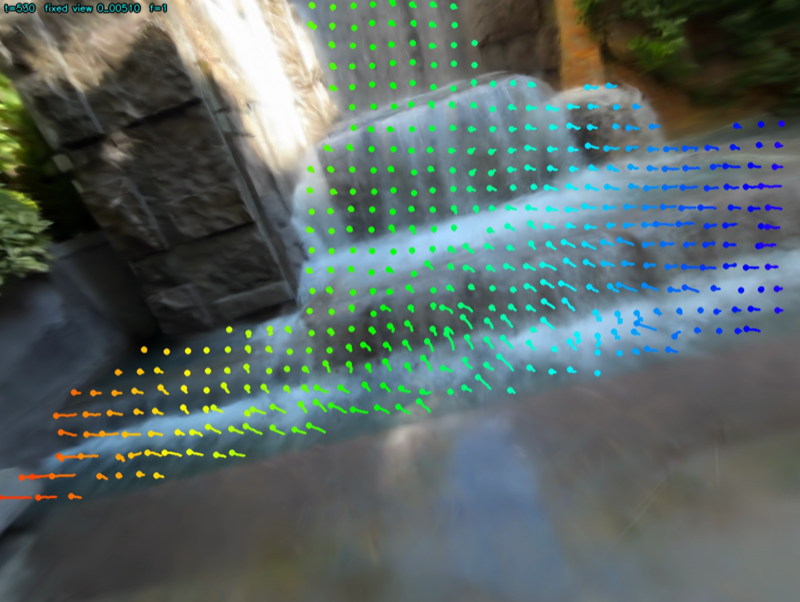}    & \tcell{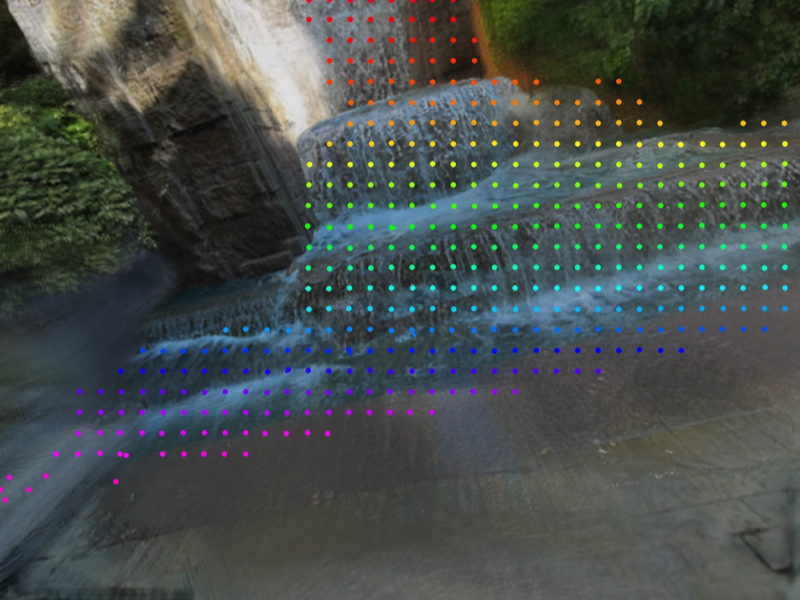}   \\[1pt]
     \tcell{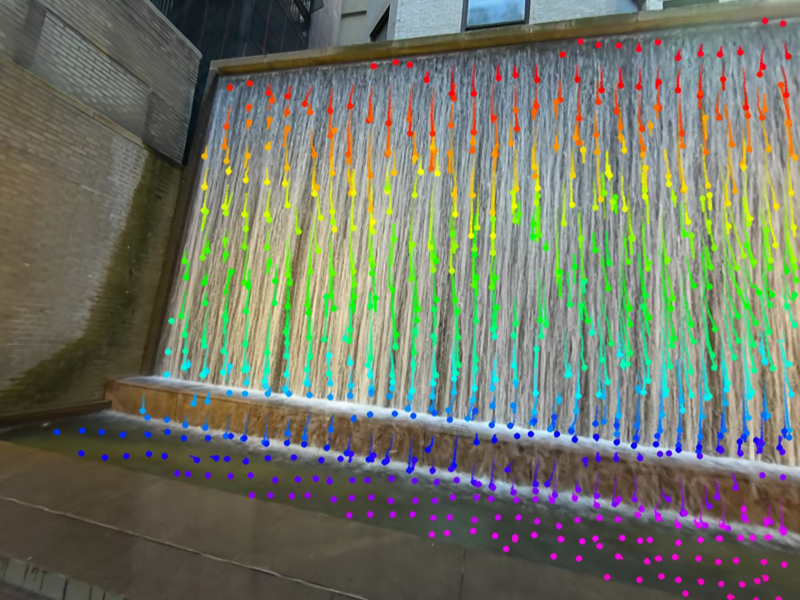} & \tcell{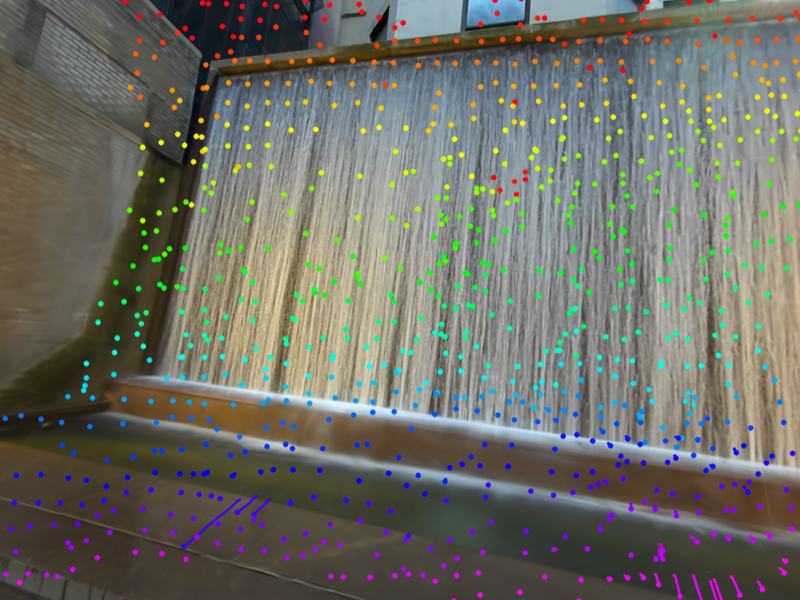} & \tcell{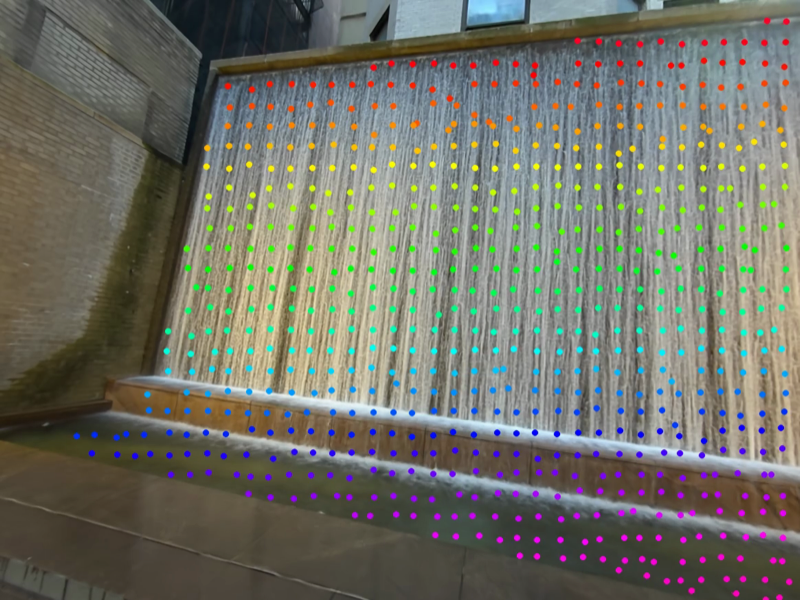} & \tcell{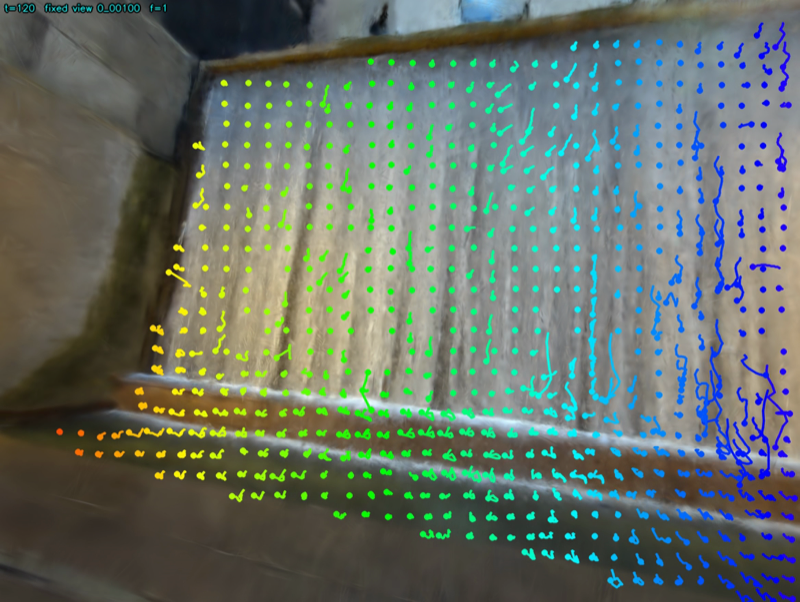}  & \tcell{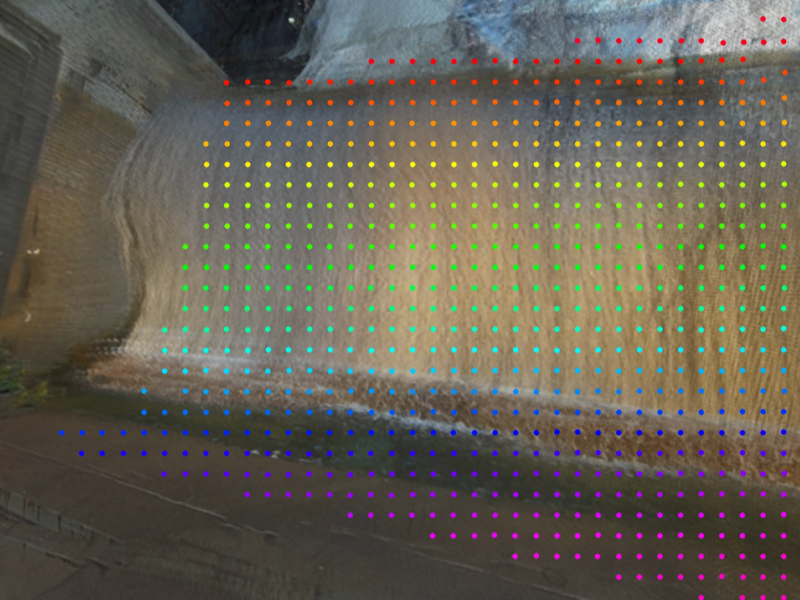} \\
   \end{tabular}}
   \vspace{-5pt}
  \caption{\textbf{Scene flow comparison.} We visualize tracks formed by canonical Gaussian Splats by iteratively querying our Eulerian motion field with residual applied, on two scenes (rows),
  against 4DGS~\cite{Wu_2024_CVPR}, AmbGS~\cite{Shih2024AmbientGS},
  MoSca~\cite{lei2024mosca} and MoVieS~\cite{lin2026movies}. For the
  baselines, which do not expose an explicit Eulerian motion field, we extract the
  corresponding tracks by querying each method's own motion representation at the same set of points and time steps. Our Eulerian motion field yields long, coherent trajectories that follow natural water flow, whereas the baselines produce short, incoherent, or
  near-static tracks.}\vspace{-10pt}
  \label{fig:flow_field}
\end{figure}

\subsection{Qualitative Comparison}

\begin{figure*}[htbp]
  \centering
 \includegraphics[width=\textwidth]{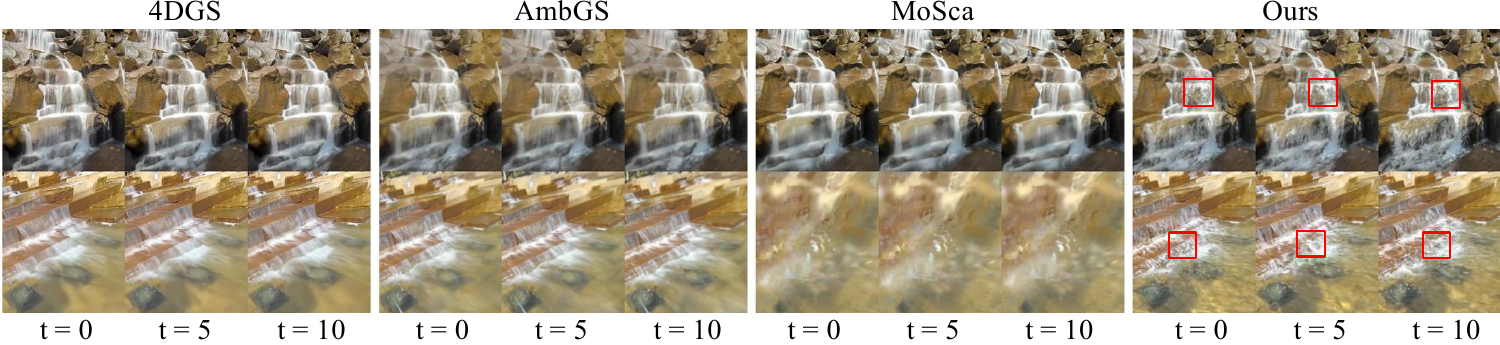}
 \vspace{-22pt}
  \caption{\textbf{Visualizing temporal progression of water.} We compare our method to the baselines at different time steps. Our method reproduces natural flow of water, while 4DGS~\cite{Wu_2024_CVPR} AmbGS~\cite{Shih2024AmbientGS} and MoSca~\cite{lei2024mosca} only demonstrate slight shift in color over time.}\vspace{-10pt}
  \label{fig:timeseries}
\end{figure*}

We show visual comparison with baselines against ground truth image in Fig.~\ref{fig:gt_comp}.
Overall, our method captures the most detail in the water regions. 
While AmbGS reconstructs most of the static part of the scene, its periodic motion model is too restrictive and cannot fit to our input video. MoSca relies strongly on image-based priors such as 2D tracking which frequently fail for water scenery. Although 4DGS employs a flexible deformation-field based motion model which can fit to motion from input images to some degree, it does not produce natural water motion. While MoVieS can reconstruct good details, it often fails on parts of the scene with poor coverage and cannot pick up the fast but subtle water motion. 
Fig.~\ref{fig:flow_field} compares the recovered scene flow against the baselines, visualizing the projected 3D tracks obtained by advecting canonical Gaussians through each method's motion representation.
We also show motion in time samples of generated fixed view video from a novel view in Fig.~\ref{fig:timeseries} compared to the baselines. It can be demonstrated that the baseline methods struggle to reconstruct highly detailed and stochastic water motion, while our method reconstructs rich and detailed motion. Please refer to the {\bf supp. material} to examine both the details and naturalness of water motion that we reconstruct with our method in comparison to baselines.

\begin{table}[t]
\centering
\setlength{\tabcolsep}{3pt}
\resizebox{\columnwidth}{!}{\small
\begin{tabular}{lcccccc}
\toprule
Cycle length & PSNR ↑ & SSIM ↑ & LPIPS ↓ & FID ↓ & KID ↓ & FVD ↓ \\
\midrule
$L=35$\phantom{w/o opacity+color deformation}  & 22.78 & 0.702 & 0.329 & 42.18 & 0.008 & 247.76 \\
$L=25$                          & \snd{22.87} & \snd{0.708} & \snd{0.321} & \best{\textbf{39.35}} & \best{\textbf{0.007}} & \snd{210.96} \\
\midrule
$\mathbf{L=15}$ \textbf{(ours)} & \best{\textbf{23.05}} & \best{\textbf{0.716}} & \best{\textbf{0.315}} & \snd{39.63} & \snd{0.007} & \best{\textbf{210.01}} \\
\bottomrule
\end{tabular}}
\vspace{-8pt}
\caption{\textbf{Cycle-length ablation.} Effect of the cycle length $L$ on
reconstruction quality. Metrics follow the same protocol as Tab.~\ref{tab:abl}.}\vspace{-5pt}
\label{tab:cycle_abl}
\end{table}

\begin{table}[t]
\centering
\setlength{\tabcolsep}{3pt}
\resizebox{\columnwidth}{!}{\small
\begin{tabular}{lcccccc}
\toprule
Method & PSNR ↑ & SSIM ↑ & LPIPS ↓ & FID ↓ & KID ↓ & FVD ↓ \\
\midrule
w/o Eulerian motion field       & \best{\textbf{23.42}} & \best{\textbf{0.733}} & \snd{0.319} & 50.86 & 0.011 & 422.66 \\
w/o flow initialization         & 22.50 & 0.704 & 0.325 & 55.44 & 0.012 & 473.29 \\
w/o non-periodic residual       & 22.97 & 0.713 & 0.321 & 47.22 & \snd{0.010} & \snd{309.37} \\
Static reconstruction           & \snd{23.11} & \snd{0.723} & 0.319 & 60.88 & 0.015 & 522.97 \\
w/o Alg. 1                       & 22.86 & 0.702 & 0.329 & \snd{47.01} & \snd{0.010} & 334.30 \\
\midrule
\textbf{Full model}             & 23.05 & 0.716 & \best{\textbf{0.315}} & \best{\textbf{39.63}} & \best{\textbf{0.007}} & \best{\textbf{210.01}} \\
\bottomrule
\end{tabular}}
\vspace{-8pt}
\caption{\textbf{Ablations of our method.} Each row removes a single component of our full model or replaces it with a static reconstruction. Best per column is highlighted in red (bold) and second best in orange.}\vspace{-15pt}
\label{tab:abl}
\end{table}

\subsection{Ablations}
\label{subsec:ablations}
\begin{figure}[t]
  \centering
  {\renewcommand{\arraystretch}{0}
   \setlength{\tabcolsep}{0.5pt}
   \newcommand{\hdr}[1]{{\fontfamily{ptm}\selectfont\scriptsize #1}}
   \newcommand{\tcell}[1]{\includegraphics[width=0.245\columnwidth]{#1}}
   \begin{tabular}{@{}cccc@{}}
     \hdr{\textbf{Full model}} & \hdr{w/o Eulerian motion field} & \hdr{w/o flow init} & \hdr{w/o residual} \\[1pt]
     \tcell{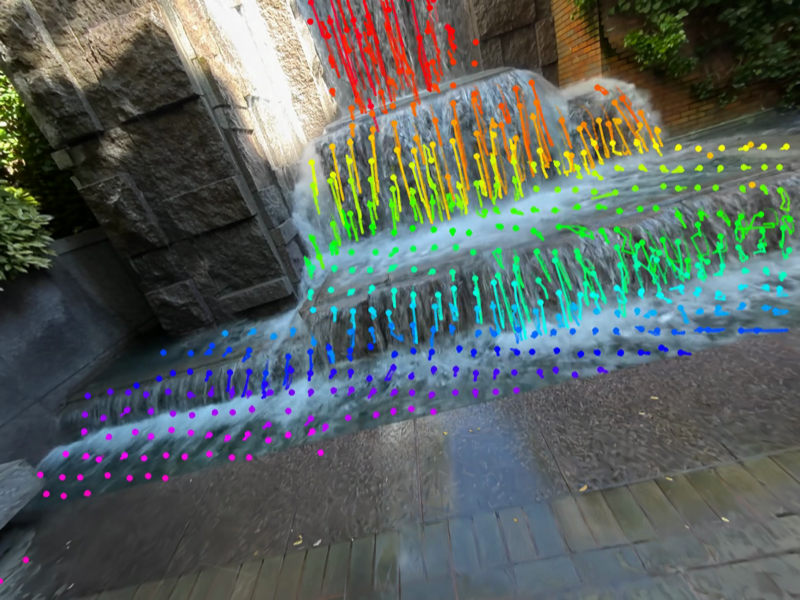}    & \tcell{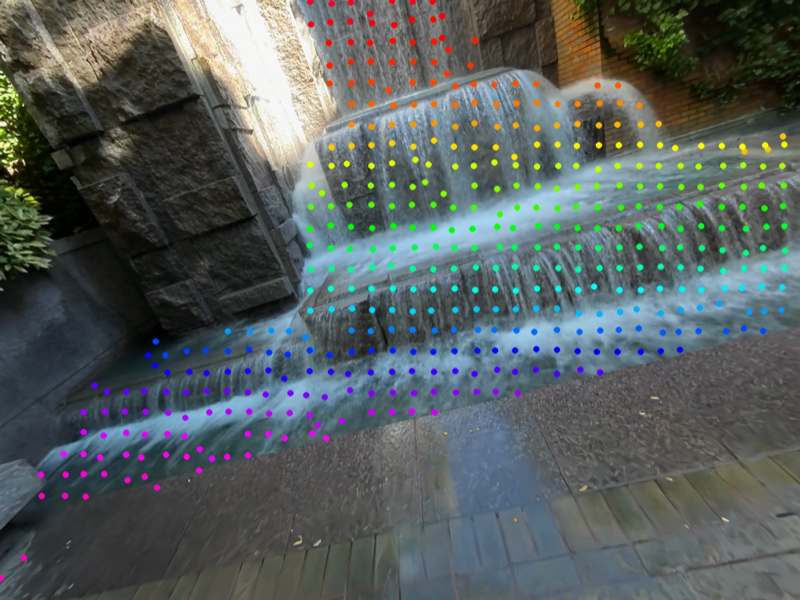}    & \tcell{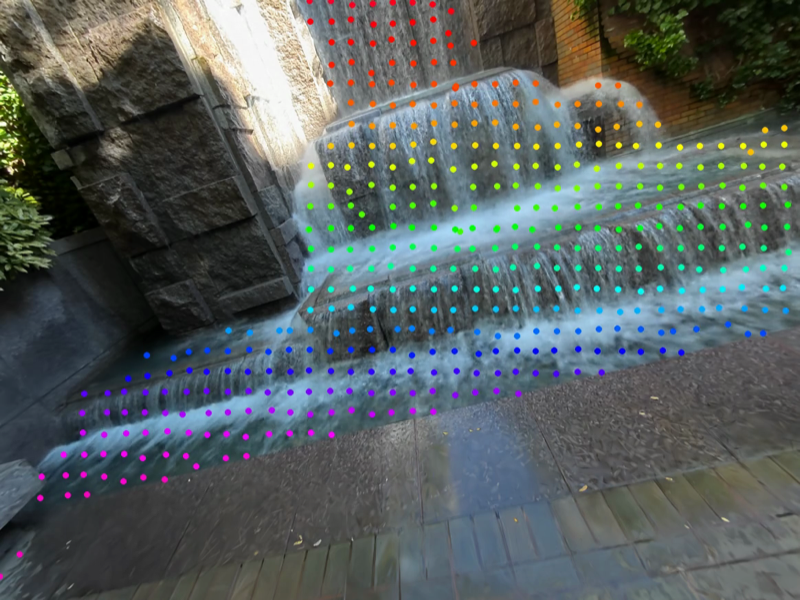}    & \tcell{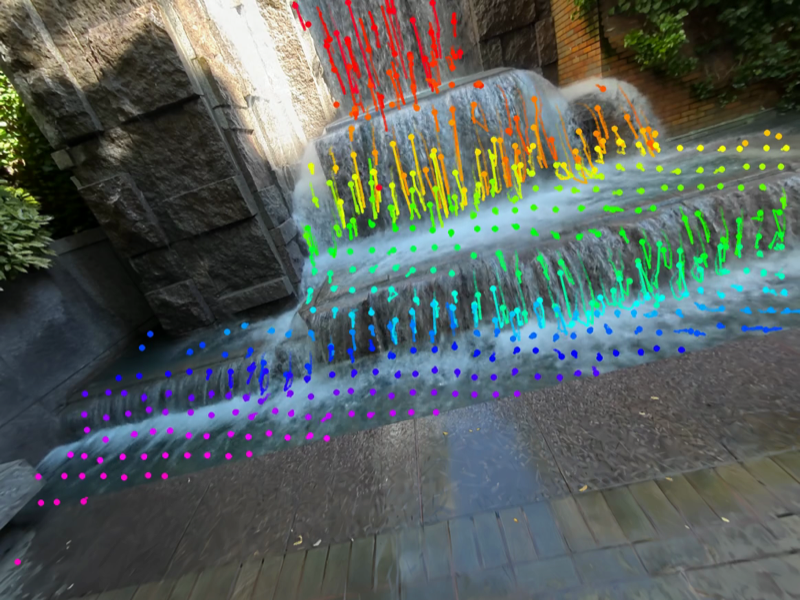}    \\[1pt]
     \tcell{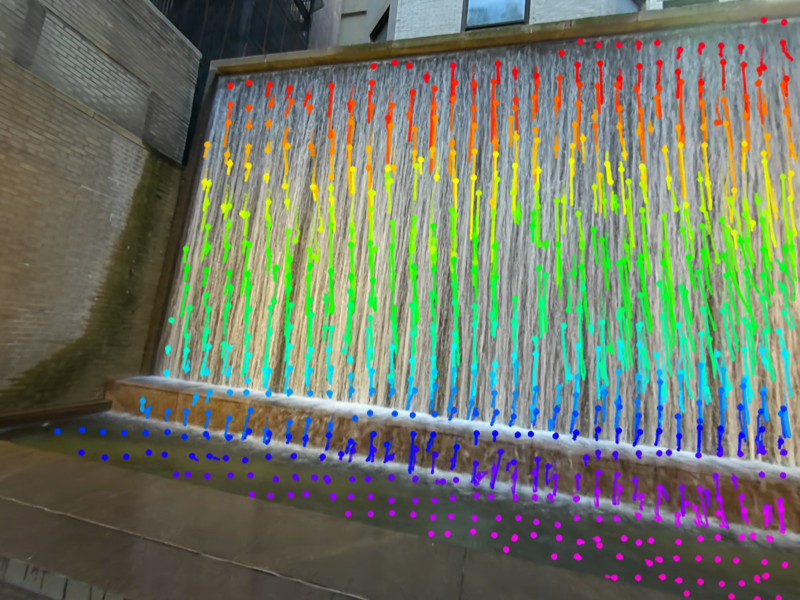}  & \tcell{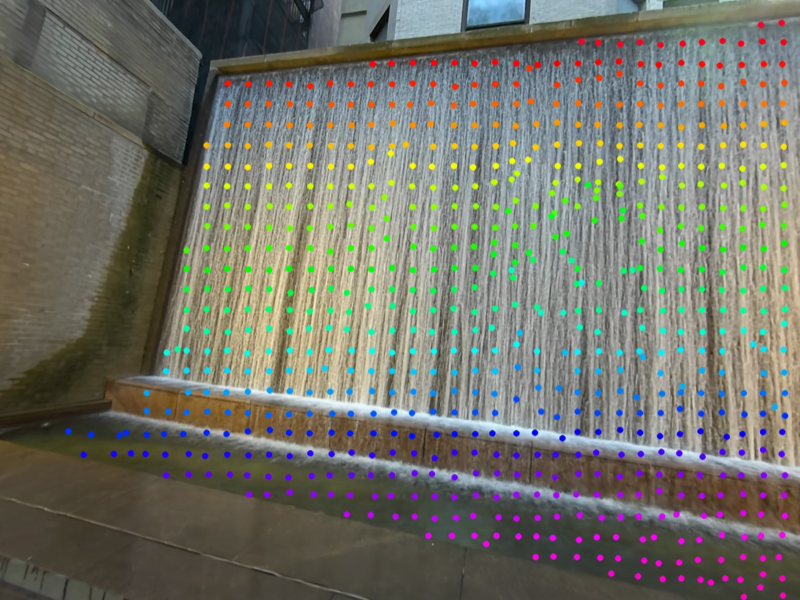}  & \tcell{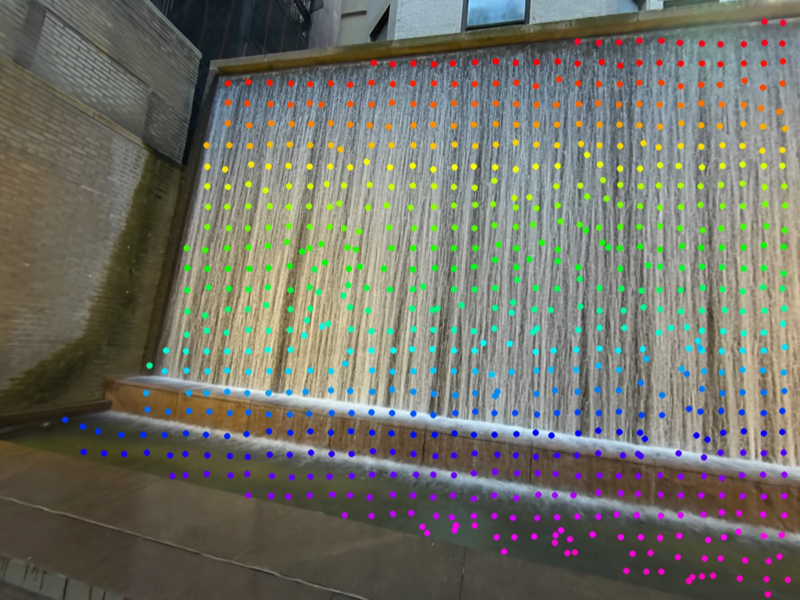}  & \tcell{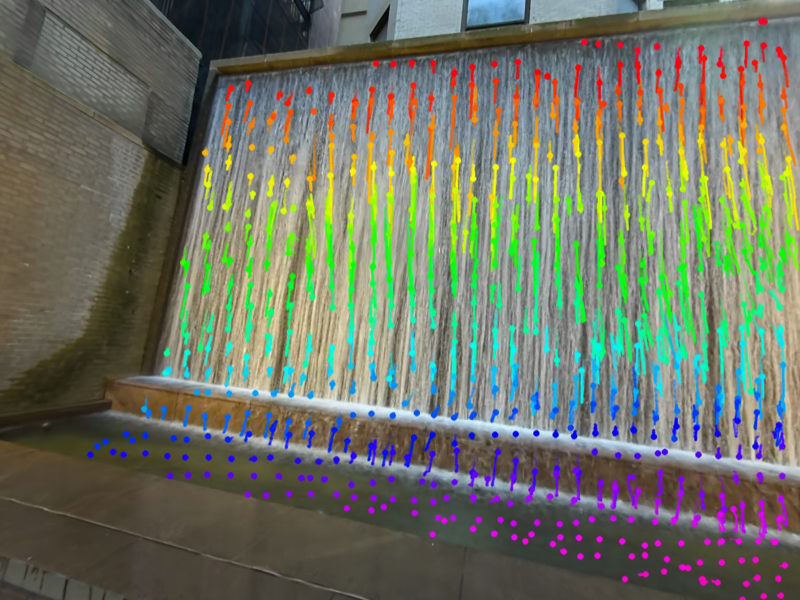}  \\
   \end{tabular}}
  \vspace{-6pt}
  \caption{\textbf{Ablation track visualization.} Tracks formed by iteratively querying each ablation variant's motion representation on two scenes, shown at the final frame of a 15-frame propagation. The full model produces coherent water-following trajectories; model with the  Eulerian motion field or  flow initialization removed barely reconstructs and  water flow motion, and removing the non-periodic residual produces less natural and stochastic tracks.}\vspace{-13pt}
  \label{fig:ablation}
\end{figure}

We perform a qualitative and quantitative ablation study to evaluate the contribution of each component in our framework; see Tab.~\ref{tab:abl} and Fig.~\ref{fig:ablation}. We conduct experiments where we remove flow initialization (Sec.~\ref{subsec:init}), so that our motion field is optimized from scratch; remove the residual deformation field that predicts non-periodic position, opacity and color offsets (Sec.~\ref{subsec:Non-periodic}); and lastly remove our Eulerian motion field, so that the motion is only represented by the residual deformation field. Tab.~\ref{tab:abl} shows that our full model achieves the best perceptual and distributional metrics compared to the models without each of the components.  Fig.~\ref{fig:ablation} shows that the full model produces coherent water-following trajectories while without the Eulerian motion field or flow initialization, only small amount of motion is reconstructed and without the non-periodic residual, reconstructed motion is less stochastic corresponding to the limited capability of our static Eulerian motion field.  Per-component reconstruction visualizations for these ablations are provided in the supplementary (Fig.~\ref{fig:abl_tracks}). This set of ablation studies highlights our insight that it is essential to leverage an Eulerian motion field to represent water motion as it allows us to benefit from the highly-effective initialization from optical flow based on nearby frames.

\begin{figure}[t]
  \centering
  \setlength{\tabcolsep}{0pt}
  \renewcommand{\arraystretch}{1.05}
  \newlength{\xtcellw}\setlength{\xtcellw}{0.18\columnwidth}
  \newlength{\xtimgh}\setlength{\xtimgh}{0.11\columnwidth}
  \newcommand{\framecell}[1]{
    \makebox[\xtcellw][c]{\includegraphics[width=\xtcellw,height=\xtimgh,keepaspectratio]{#1}}
  }
  \newcommand{\xtimg}[1]{
    \makebox[\xtcellw][c]{
    \begin{tikzpicture}[inner sep=0pt,outer sep=0pt]
      \node[anchor=south west,inner sep=0pt] (I) {\includegraphics[width=\xtcellw,height=\xtimgh,keepaspectratio]{#1}};
      \begin{scope}[overlay]
        \draw[->,line width=0.4pt,>=stealth] ([yshift=2pt]I.north west) -- ++(6pt,0)
          node[right=1pt,font=\tiny] {$x$};
        \draw[->,line width=0.4pt,>=stealth] ([xshift=-2pt]I.north west) -- ++(0,-6pt)
          node[below=1pt,font=\tiny] {$t$};
      \end{scope}
    \end{tikzpicture}}
  }
  \resizebox{\columnwidth}{!}{
  \begin{tabular}{@{}c@{\hspace{-3pt}}c@{\hspace{-3pt}}c@{\hspace{0pt}}c@{}}
    \multirow{2}{*}{\framecell{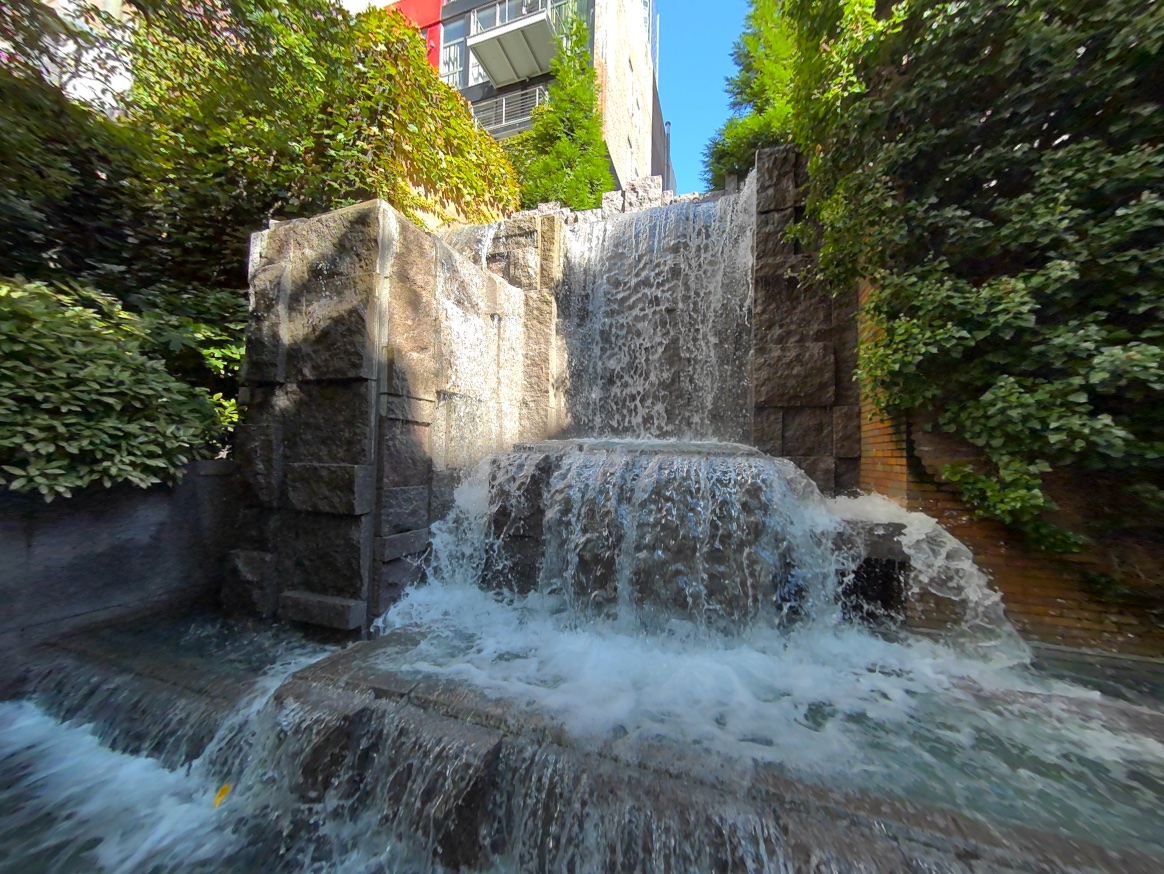}} &
    \rotatebox{90}{\tiny Ours} &
    \framecell{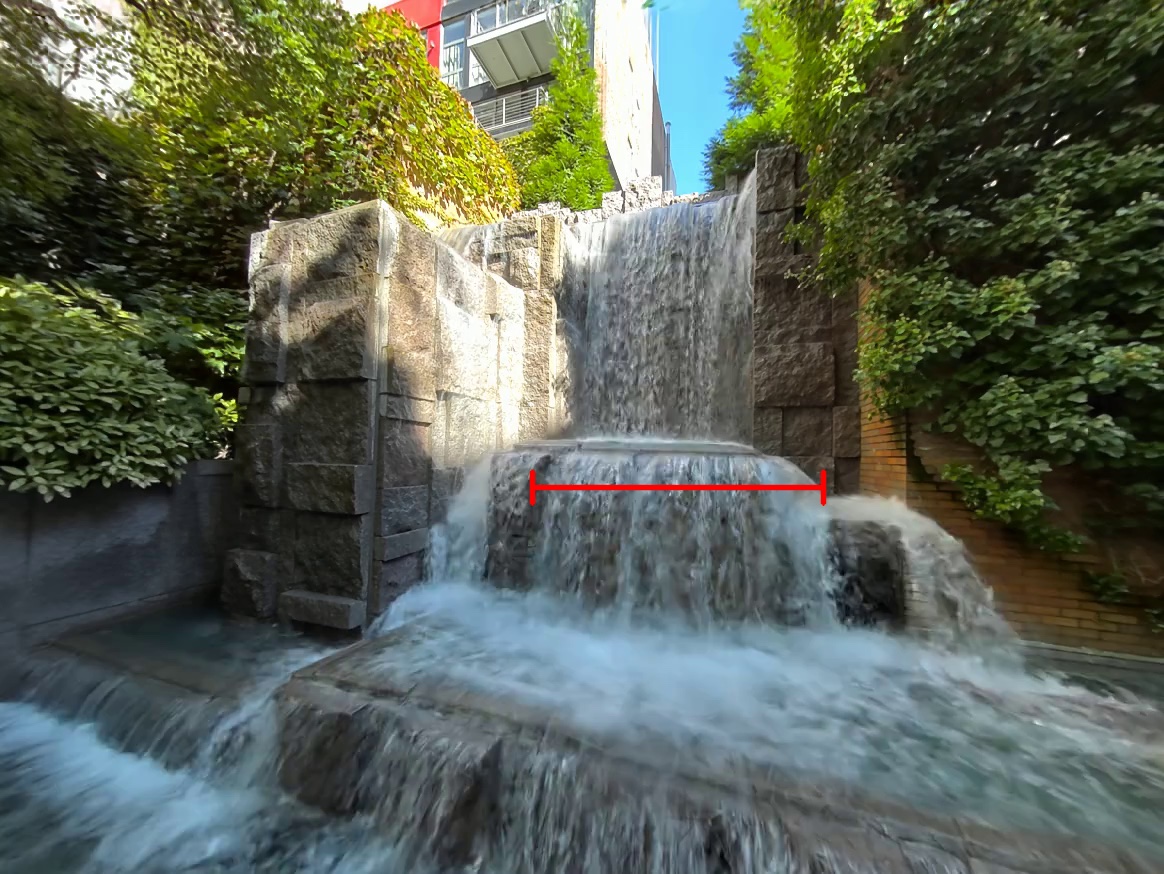} &
    \xtimg{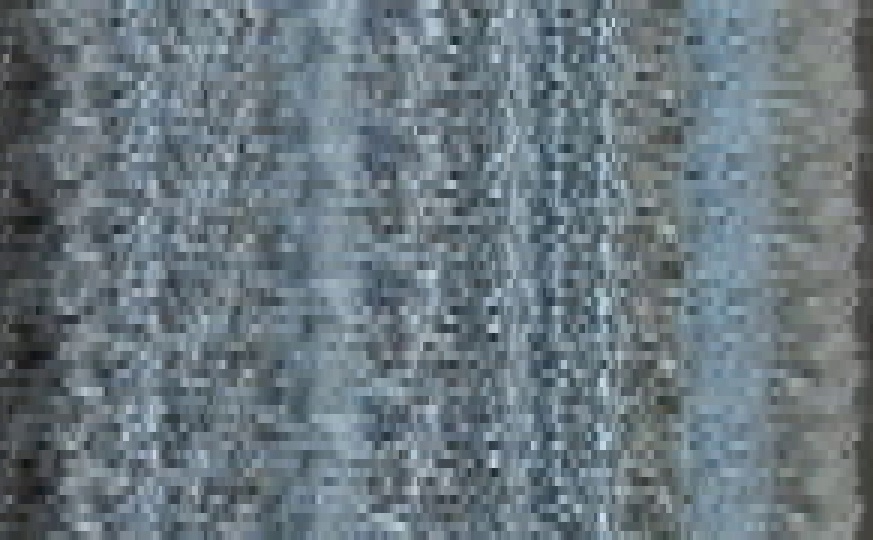} \\
    & \rotatebox{90}{\tiny w/o Alg. 1} &
    \framecell{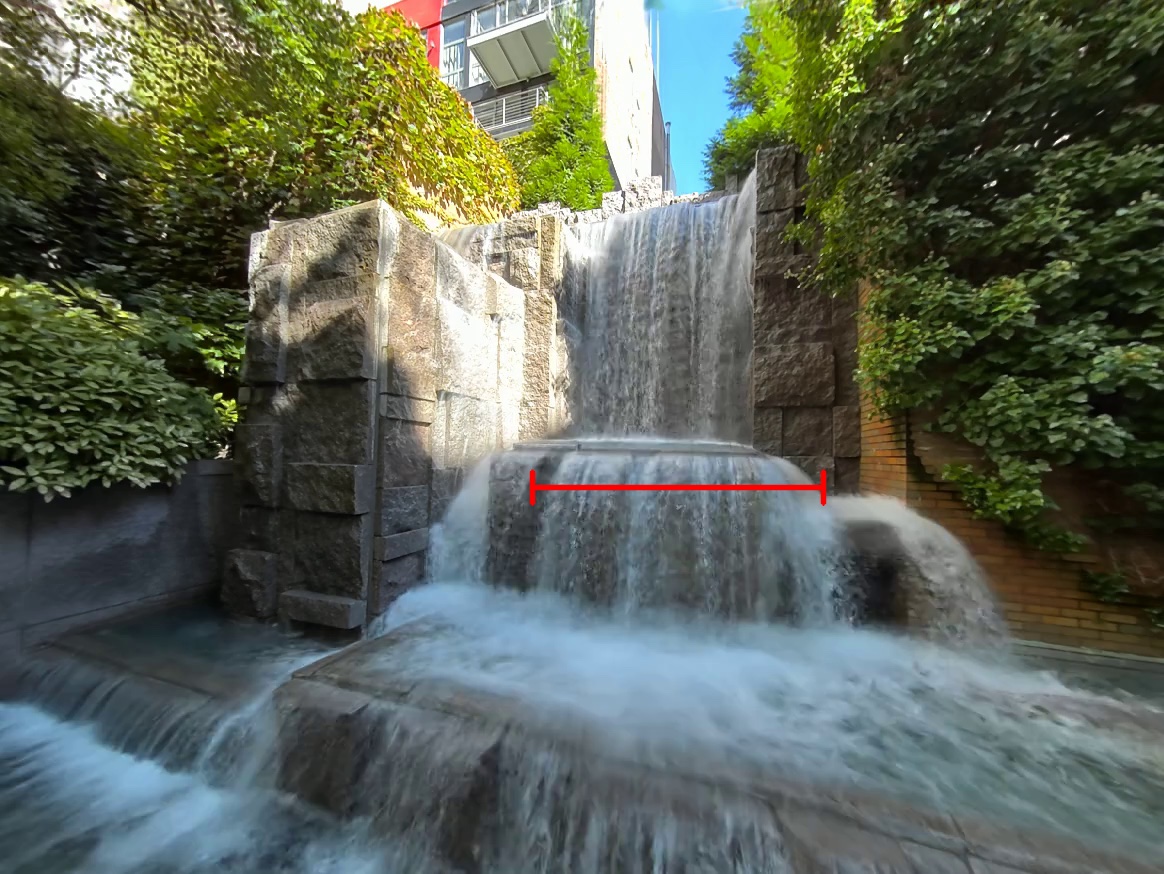} &
    \xtimg{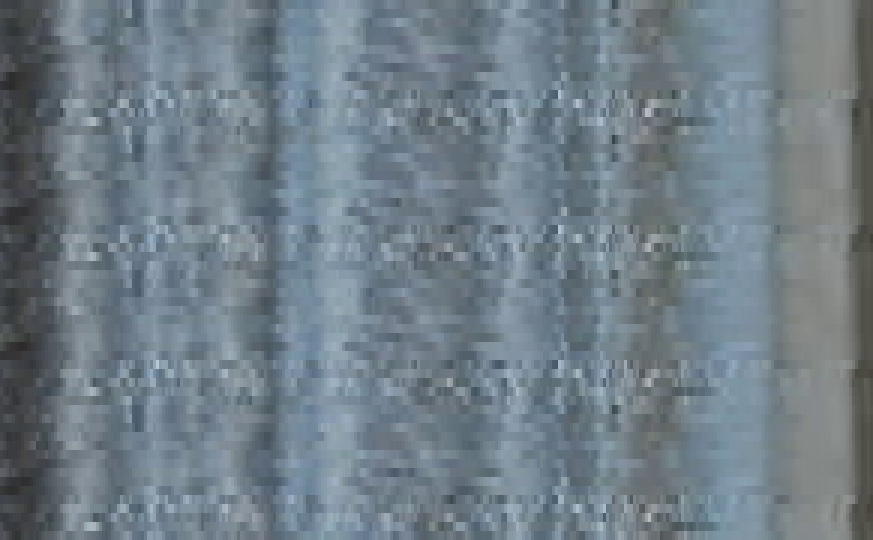} \\
    \multirow{2}{*}{\framecell{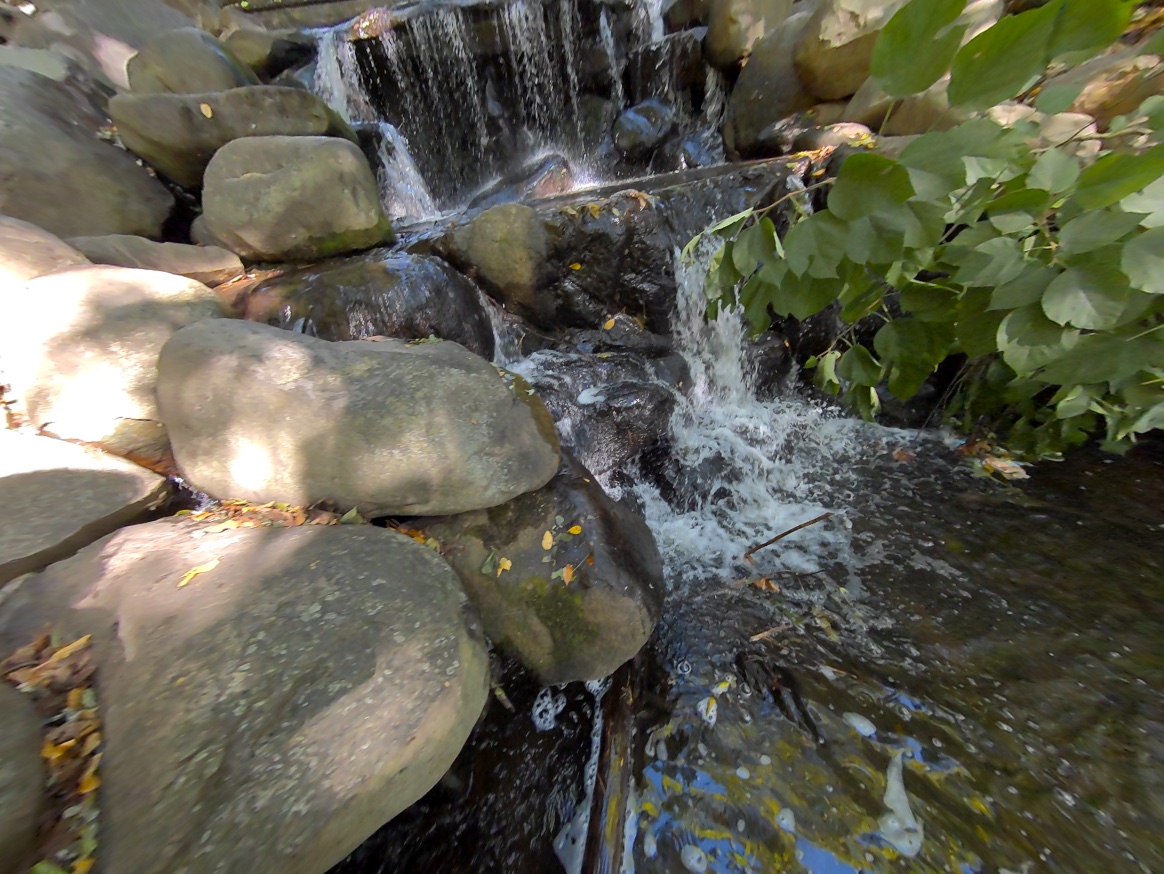}} &
    \rotatebox{90}{\tiny Ours} &
    \framecell{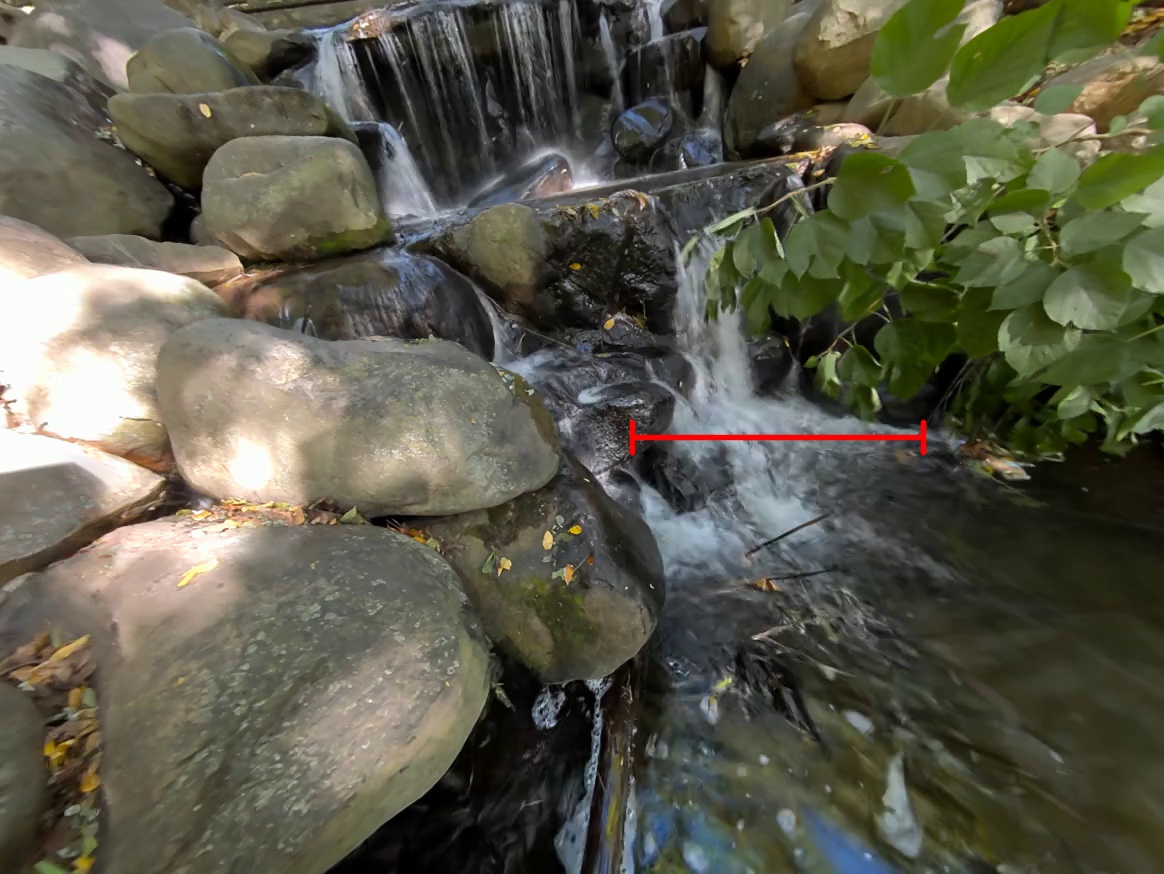} &
    \xtimg{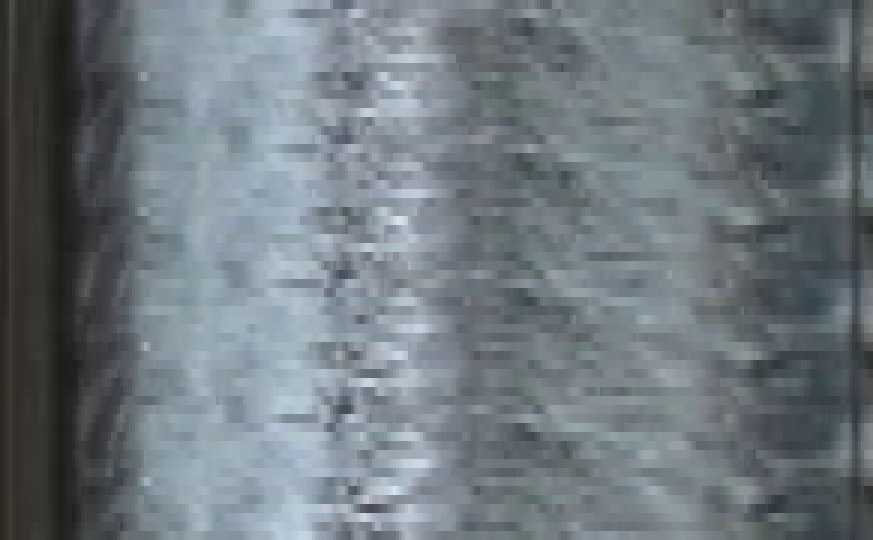} \\
    & \rotatebox{90}{\tiny w/o Alg. 1} &
    \framecell{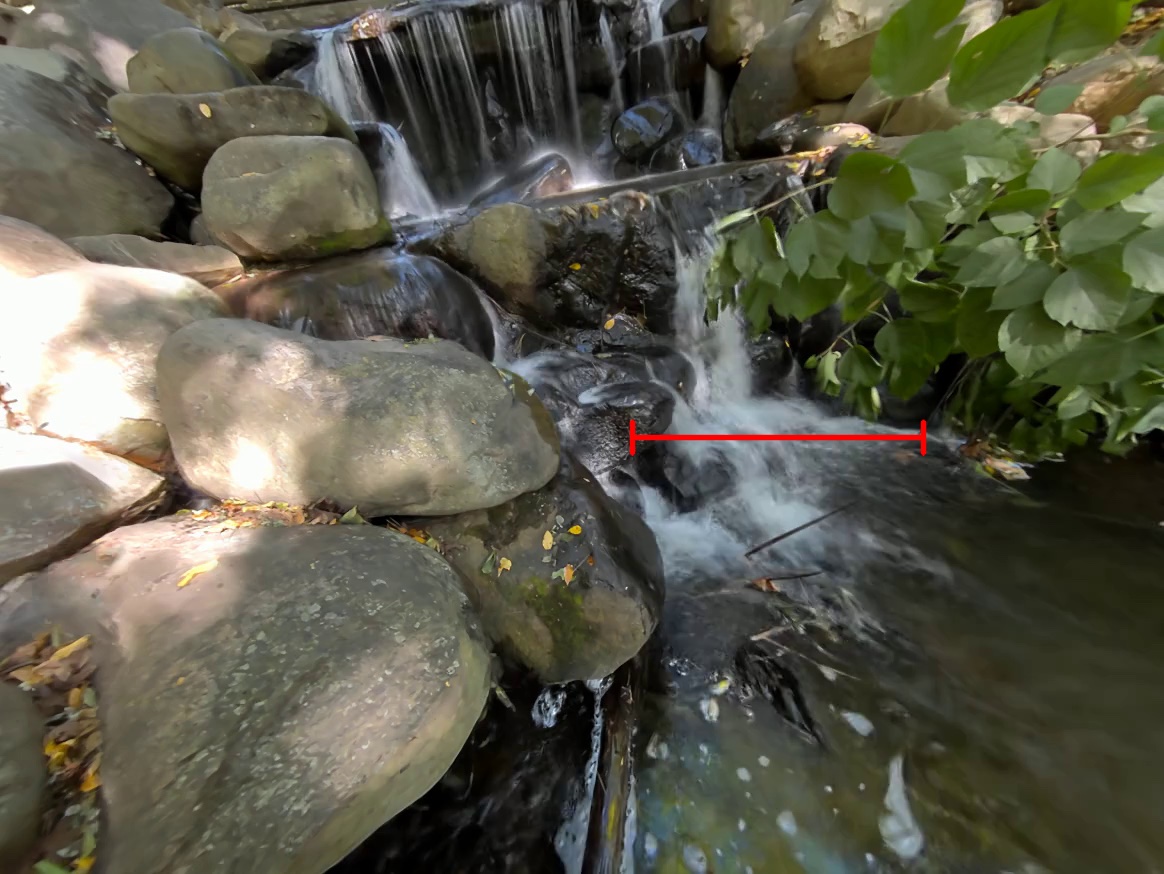} &
    \xtimg{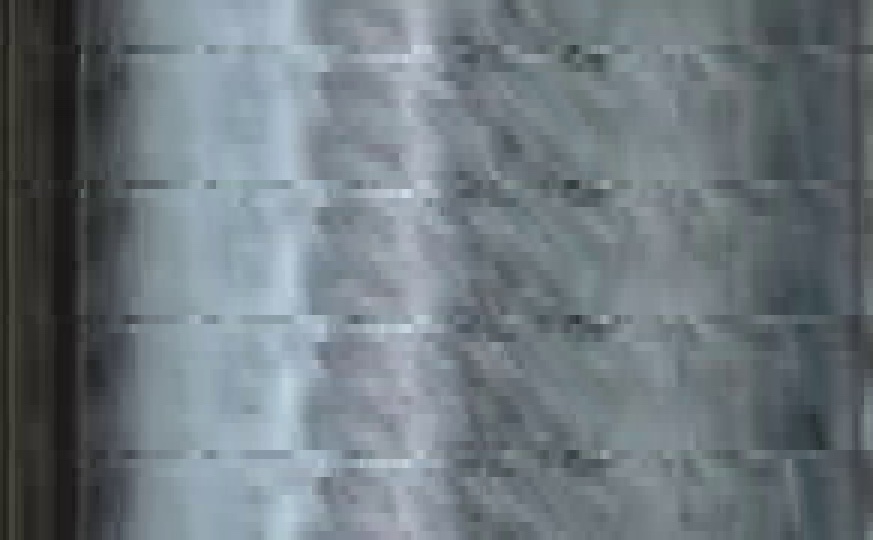} \\[-6pt]
    {\tiny (a) GT} & &
    {\tiny (b) sampled row} &
    {\tiny (c) XT slice} \\
  \end{tabular}}
  \vspace{-10pt}
  \caption{\textbf{Random start time prevents "looping artifacts"}
    For each scene we fix the camera, advance time for 60 frames (4 periods with 15 frames in each period), and stack a single
    horizontal scanline (red, in the sampled-row frames (b)) into an
    $x$--$t$ image (c); time runs downward. GT (a) is the reference. Without random
    start times (bottom row per scene, ``w/o Alg.~1'') the Gaussians pulse as one body, resulting in the slice being crossed by \emph{horizontal} bands that span the full width at a single instant at loop boundaries. With random
    start times (top row per scene, ``Ours'') water particles spread stochastically across loops. Per-frame
    metrics barely register this ($-0.19$\,dB PSNR), while FVD rises significantly.} \vspace{-10pt}
  \label{fig:xt-slice}
\end{figure}

We further discuss the \emph{w/o Alg. 1} variant in Tab.~\ref{tab:abl}, which removes the random per-Gaussian start times in Alg.~\ref{alg1} (all Gaussians reborn at $t=0$). It performs much worse than the full model on the distributional and video-quality metrics (FID/KID/FVD), confirming that staggering rebirths across the cycle helps hide the discontinuity introduced by looping. Fig.~\ref{fig:xt-slice} visualizes this effect via spatiotemporal $x$--$t$ slices on two scenes, where without random per-Gaussian start times there is an obvious discontinuity across loops. 

\begin{figure}[t]
  \centering
  {\renewcommand{\arraystretch}{0}
   \setlength{\tabcolsep}{1pt}
   \newcommand{\hdr}[1]{{\fontfamily{ptm}\selectfont\small #1}}
   \newcommand{\cimg}[1]{\includegraphics[width=0.295\columnwidth,viewport=582 437 1164 874,clip]{#1}}
   \begin{tabular}{@{}c@{\hspace{2pt}}ccc@{}}
   & \hdr{$L=15$} & \hdr{$L=25$} & \hdr{$L=35$} \\[2pt]
   \hdr{\raisebox{15pt}{\rotatebox{90}{$t=1$}}} & \cimg{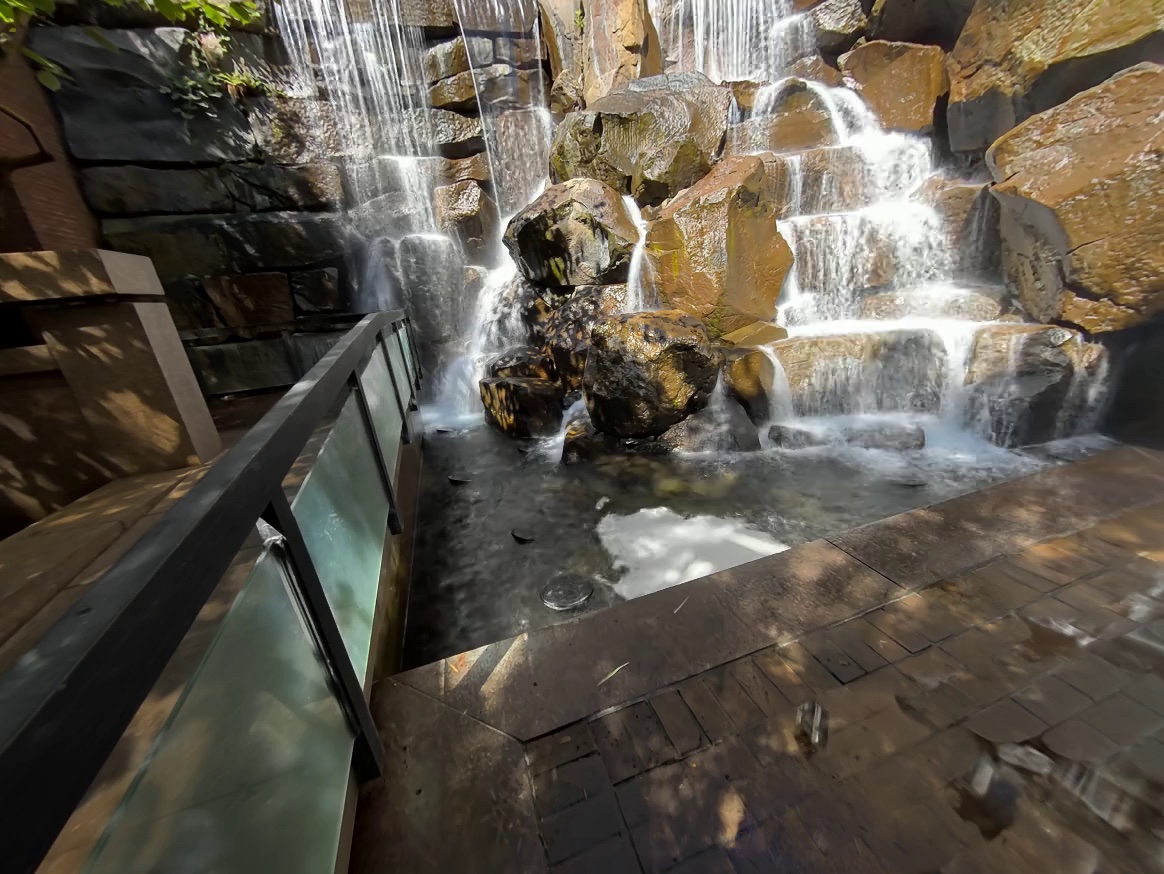} & \cimg{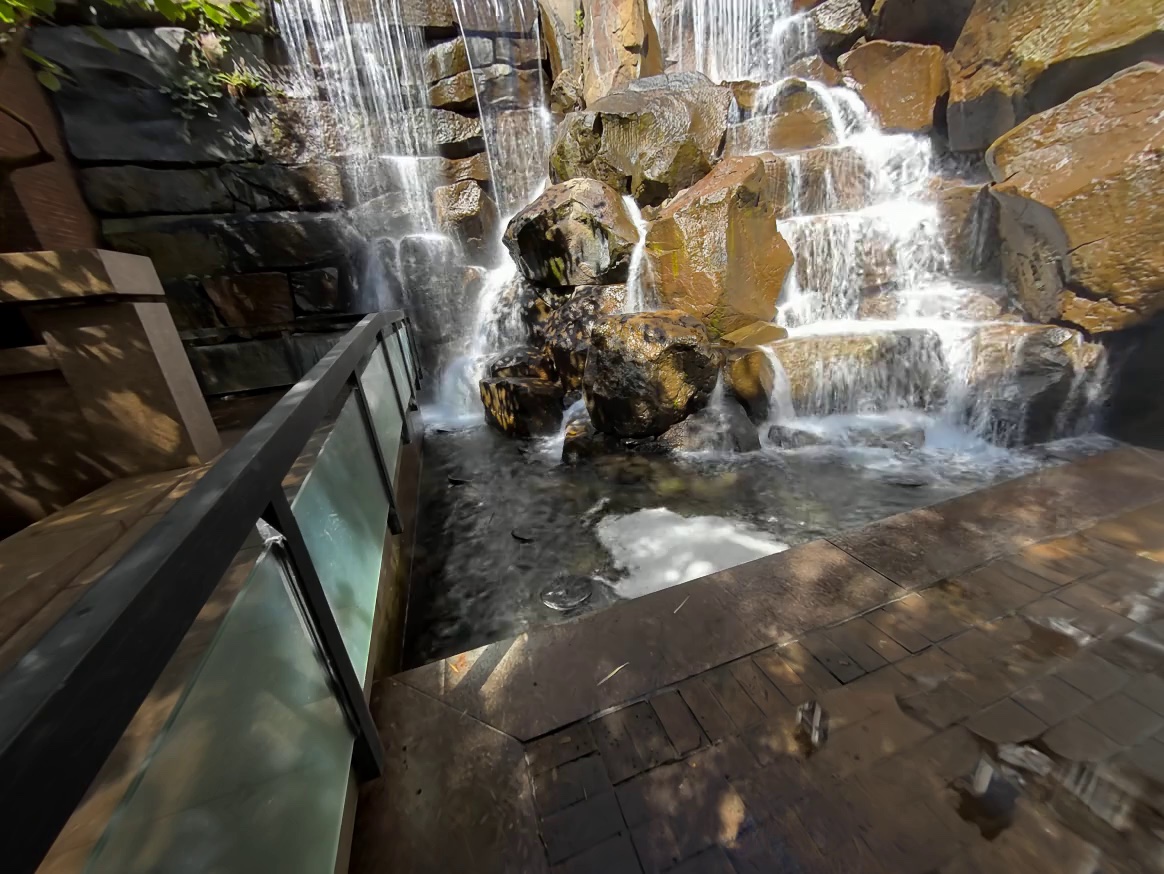} & \cimg{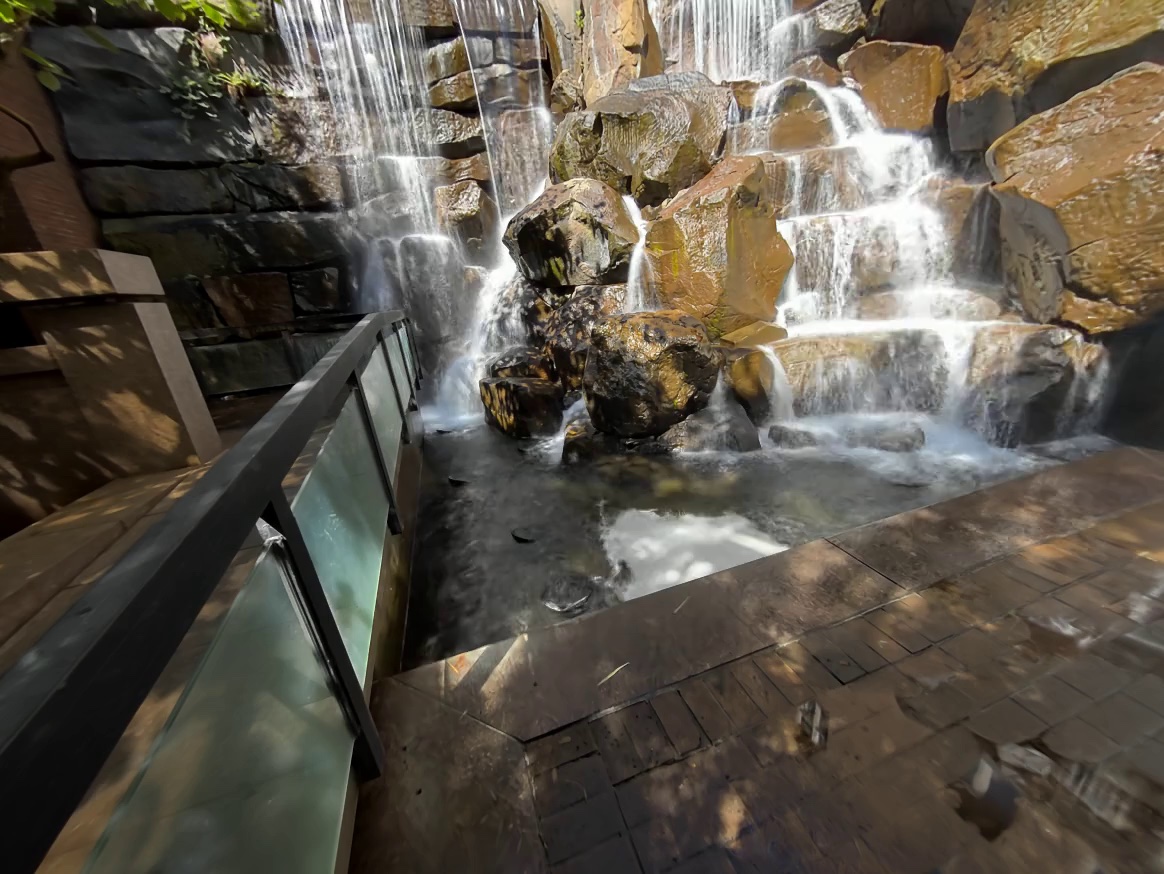} \\[2pt]
   \hdr{\raisebox{15pt}{\rotatebox{90}{$t=5$}}} & \cimg{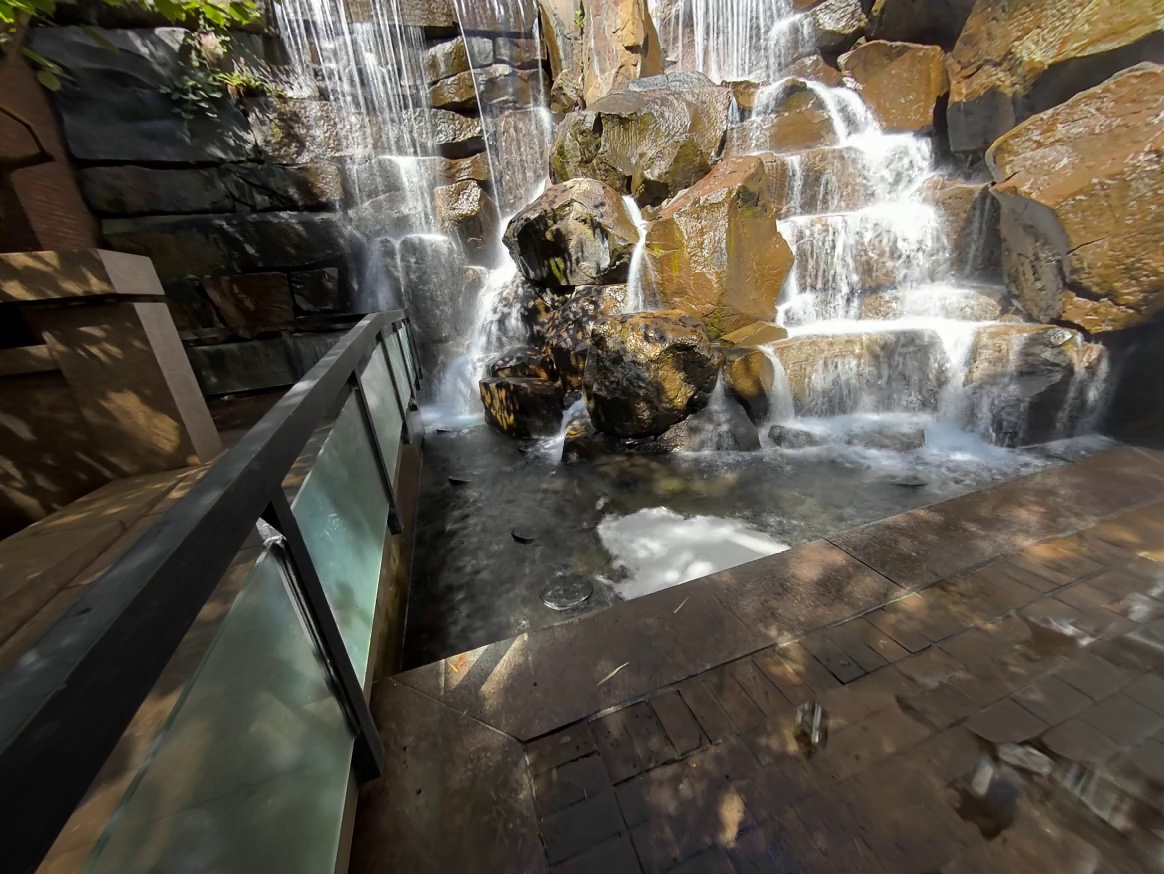} & \cimg{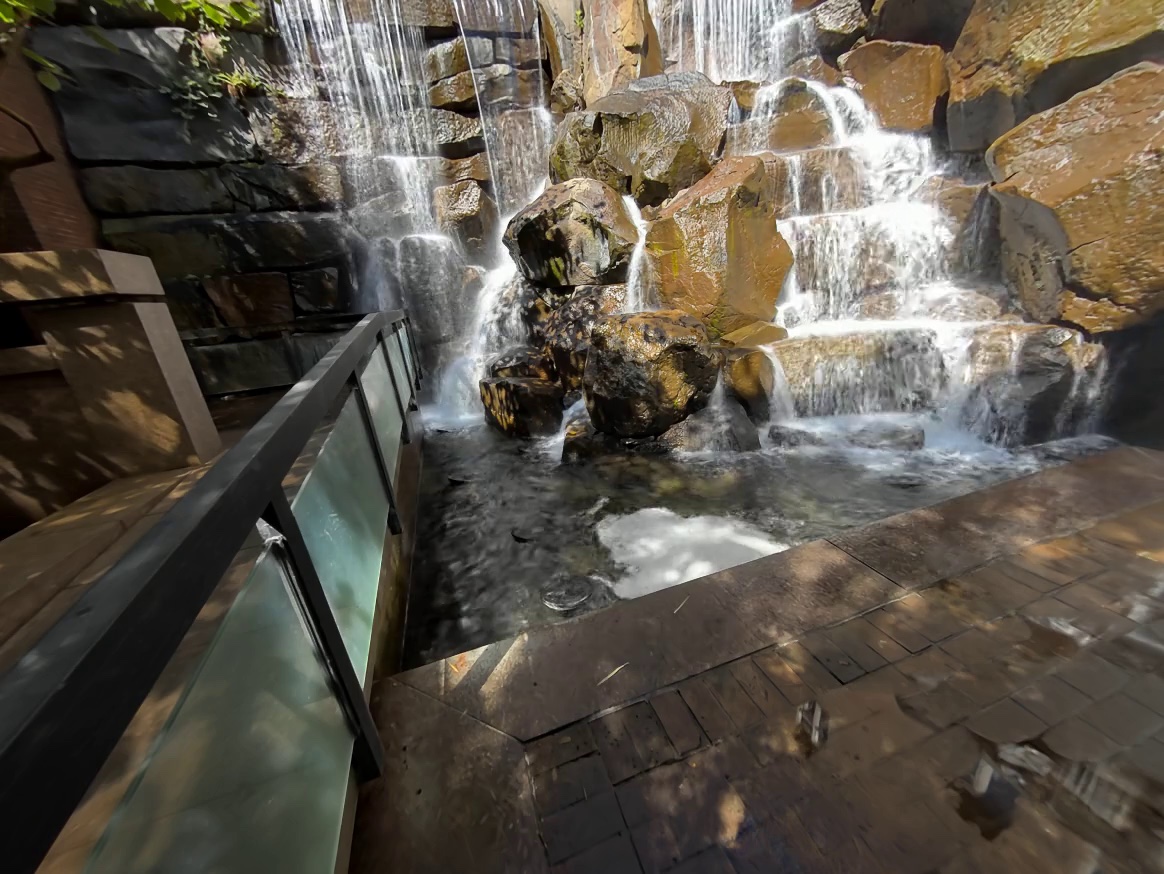} & \cimg{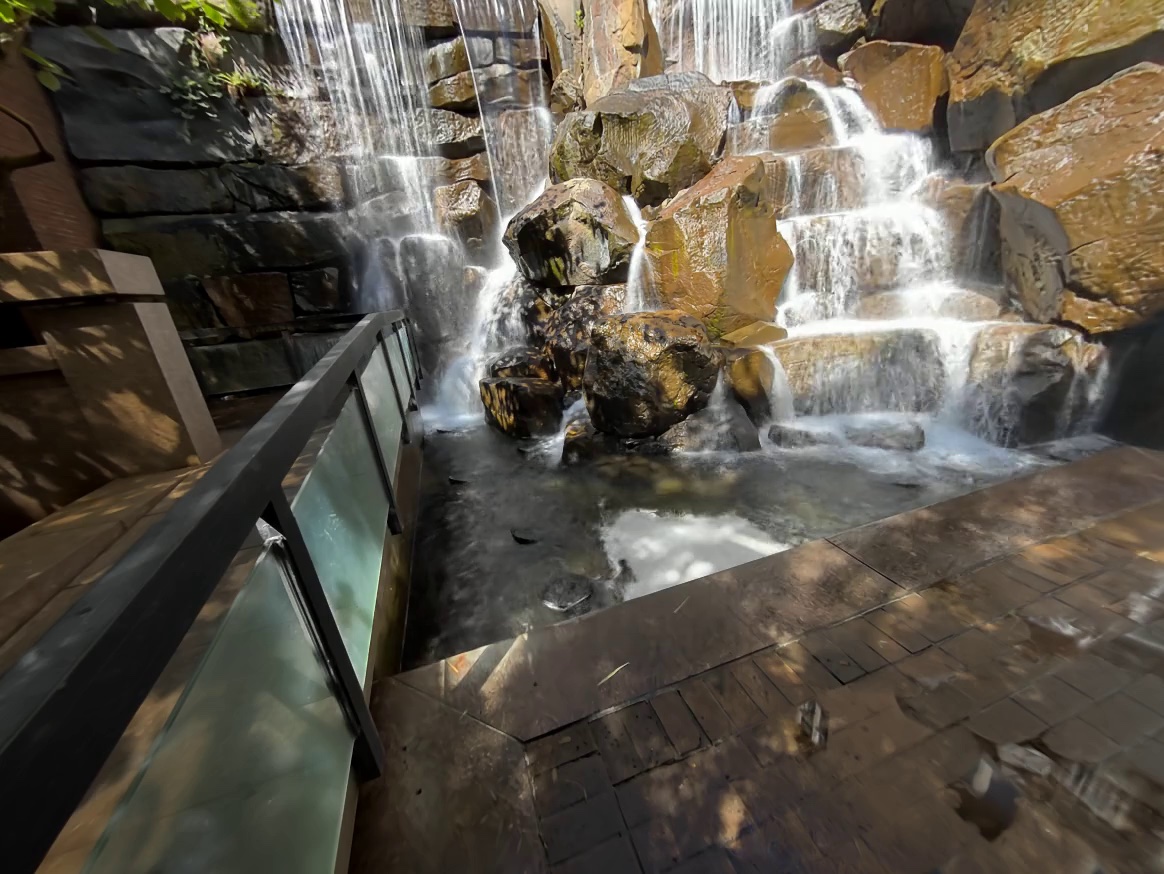} \\
   \end{tabular}}
   \vspace{-5pt}
  \caption{\textbf{Effect of changing cycle length $L=(15, 25, 35)$.} Columns vary the cycle length $L$; rows show two time steps in the loop. It can be observed that L=25 produces better dynamics (more variation in the waterfall states) compared to L=15, while L=35 results in blurrier overall reconstruction due to worse convergence. }\vspace{-10pt}
  \label{fig:n_ablation}
\end{figure}

We additionally ablate the cycle length $L$, which controls the temporal span of a single loop. Fig.~\ref{fig:n_ablation} compares three settings qualitatively and Tab.~\ref{tab:cycle_abl} reports the corresponding metrics. While using larger L allows representing more diverse motion, it also leads to poorer convergence hence lower overall reconstruction quality. We adopt $L=15$, which best balances the range of motion within a cycle against reconstruction sharpness.

\section{Discussion}
\vspace{-5pt}

\begin{figure}[t]
  \centering
  \includegraphics[width=0.8\columnwidth]{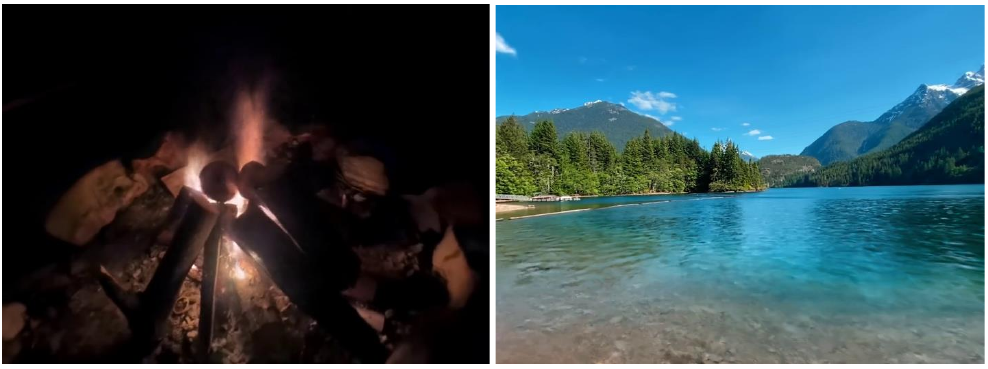}
  \vspace{-5pt}
  \caption{\textbf{Extensions.} With the help of SFM tools such as MegaSam~\cite{li2025megasam}, our method can be extended to reconstruct motion of water surfaces, \eg, lakes. We also demonstrate that our Eulerian motion field can represent other media such as fire by extending it to a time-varying model.}\vspace{-15pt}
  \label{fig:limitation}
\end{figure}

\paragraph{Limitations and future work}
While our proposed pipeline reconstructs motion of water sceneries realistically, it can fail for large water surfaces with reflections, which are common in rivers, lakes, and oceans. The reason is that, without any explicit modeling of reflections, the Gaussian Splatting algorithm reconstructs fake geometry at the bottom of the water surface to reproduce the appearance changes caused by reflections, and as a result, the water surface is reconstructed as a hazy volume with incorrect depth. Since our model requires unprojecting optical flow to 3D using depth from the static reconstruction to initialize our Eulerian motion field, incorrect depth of the water surface would make it fail. We hope to leverage monocular depth-based SFM tools such as MegaSam\cite{li2025megasam} as well as physics-based modeling of reflections such as~\cite{gao2024planar} combined with our method to extend our pipeline to more diverse water scenes. We show an example of an early attempt of reconstructing a lake scene in Fig. \ref{fig:limitation}. Additionally, the constant assumption of our current Eulerian motion field limits its capability of representing stochastic and time-varying motion. We hope to extend our Eulerian motion field to be able to model time-varying motion by initializing a time-varying Eulerian motion field from a static one, and we show preliminary results of this on reconstructing fire in Fig. \ref{fig:limitation}.
\paragraph{Conclusion}
\vspace{-13pt}
This paper proposes a novel pipeline to reconstruct and animate dynamics of water sceneries from a casual monocular video by learning a static Eulerian motion field that advects Gaussian Splats in a loopable representation. Experiments verify the effectiveness of our approach in reconstructing such motion over the baselines through qualitative and quantitative comparisons.

\clearpage
\appendix

\setcounter{section}{0}
\renewcommand{\thesection}{\Alph{section}}
\setcounter{figure}{0}
\renewcommand{\thefigure}{\Alph{section}\arabic{figure}}
\setcounter{table}{0}
\renewcommand{\thetable}{\Alph{section}\arabic{table}}

\twocolumn[{
  \renewcommand\twocolumn[1][]{#1}
  \begin{center}
    {\Large\bfseries Supplementary Material\par}
    \vspace{0.5em}
    {\large Eulerian Motion Reconstruction for Water Scenery\par}
  \end{center}
  \vspace{1.0em}
}]

\section{Overview}
\label{sec:supp_overview}

This supplementary material provides additional details and results that complement the main paper:
\begin{itemize}
  \item Additional implementation details (Sec.~\ref{sec:supp_impl}), covering experimental setup (Sec.~\ref{sec:supp_expt}), hyperparameters and architecture (Sec.~\ref{sec:supp_hyper}), and dataset capture details (Sec.~\ref{sec:supp_data}).
  \item Additional ablations and results (Sec.~\ref{sec:supp_extra}), including per-component reconstruction visualisations (Sec.~\ref{sec:supp_abl_tracks}).
  \item User study interface and instructions (Sec.~\ref{sec:supp_user}).
\end{itemize}
We strongly encourage the reader to view the accompanying \textbf{video results} in the supplementary webpage, which best convey the reconstructed water dynamics.

\section{Additional Implementation Details}
\label{sec:supp_impl}
Here we provide additional implementation details that complement the main paper. We elaborate on experimental details (Sec.~\ref{sec:supp_expt}), hyperparameters and architecture (Sec.~\ref{sec:supp_hyper}), and dataset capture details (Sec.~\ref{sec:supp_data}).

\subsection{Experiment Details}
\label{sec:supp_expt}
\paragraph{Water Segmentation Mask} We use SAM 3~\cite{sam3} to acquire per-frame water segmentation masks.
 Then along static Gaussian Splat training we also learn per-Gaussian dynamic probability from the input water segmentation masks by rendering the probability into image along with color, and supervised using the segmentation masks with binary cross-entropy loss;
 the per-Gaussian dynamic probability is thresholded to identify dynamic Gaussians. We use a threshold of 0.2.
 The dynamic Gaussians are input into the Eulerian motion field $\mathbf{V}$ and residual deformation field $\mathbf{D}$ during training and animation while the static Gaussians are not.
 Dynamic and static Gaussians are concatenated together for rendering the output image. We continue refining the dynamic mask during the dynamic training stage but update their values every 10 steps.

\paragraph{Animation run time.}
Our looping formulation lets us reuse per-cycle propagation across all frames of a cycle at animation time, so rendering with Alg. 2 is significantly faster than with the training-time Alg. 1. Tab.~\ref{tab:runtime} reports wall-clock timings on a single A6000 GPU, both for the pure deformation pass and end-to-end (deformation + rasterization).

\begin{table}[t]
\centering
\setlength{\tabcolsep}{4pt}
\resizebox{\columnwidth}{!}{\small
\begin{tabular}{lccc}
\toprule
 & Alg. 1 & Alg. 2 & Speedup \\
\midrule
\multicolumn{4}{l}{\emph{15 frames ($L=15$, 1 cycle)}} \\
Deformation only            & 0.673\,s (44.88\,ms/frame)  & 0.224\,s (14.93\,ms/frame) & $3.01\times$ \\
End-to-end (incl.\ raster)  & 0.719\,s (47.95\,ms/frame)  & 0.270\,s (17.98\,ms/frame) & $2.67\times$ \\
\midrule
\multicolumn{4}{l}{\emph{450 frames ($L=15$, 30 cycles)}} \\
Deformation only            & 23.483\,s (52.19\,ms/frame) & 5.605\,s (12.46\,ms/frame) & $4.19\times$ \\
End-to-end (incl.\ raster)  & 28.232\,s (62.74\,ms/frame) & 7.072\,s (15.72\,ms/frame) & $3.99\times$ \\
\bottomrule
\end{tabular}}
\caption{\textbf{Animation run time.} Wall-clock time to render 15 frames ($L=15$, 1 cycle) and 450 frames ($L=15$, 30 cycles), for both the pure deformation pass and end-to-end (deformation + rasterization), using the training-time propagation (Alg. 1) versus our inference-time propagation (Alg. 2). Reusing per-cycle propagations across all frames of a cycle yields a $\sim\!3$--$4\times$ speedup, with a larger gain as the number of cycles grows.}
\label{tab:runtime}
\end{table}

\subsection{Hyperparameters and Architecture}
\label{sec:supp_hyper}

\paragraph{Eulerian motion field $\mathbf{V}$ architecture.}
It's a triplane with resolution of 128 per axis and 12 feature channels.
 The queried features are decoded to a 3-D velocity vector using a MLP with width of 64, depth of 2 and a final linear projection.

\paragraph{Residual field $\mathbf{D}$ architecture.}
It's a triplane with resolution of 128 per axis and 16 feature channels.
 The queried features are decoded to a 52-dimensional vector ( position, opacity, color residuals ) using a MLP with width of 128, depth of 2, and a  final linear projection. 
 In addition to positional features, the MLP takes the positional-encoded cycle index $t$ and cycle number $p$ concatenated with the positional features.

\paragraph{Optimization}
We use weights of 0.8, 0.2, and 0.001 for $L_1$, SSIM, and  $L_2$ regularization losses on the respective offsets. 
Additionally, we use a weight of 0.5 for the binary cross-entropy loss on the per-Gaussian dynamic probability.
It takes about 8 hours to train the model on each scene on 8 A6000 GPUs.

\subsection{Dataset Details}
\label{sec:supp_data}
\begin{figure}[t]
  \centering
  \setlength{\tabcolsep}{1pt}
  \begin{subfigure}[t]{0.485\columnwidth}
    \centering
    \includegraphics[width=\linewidth]{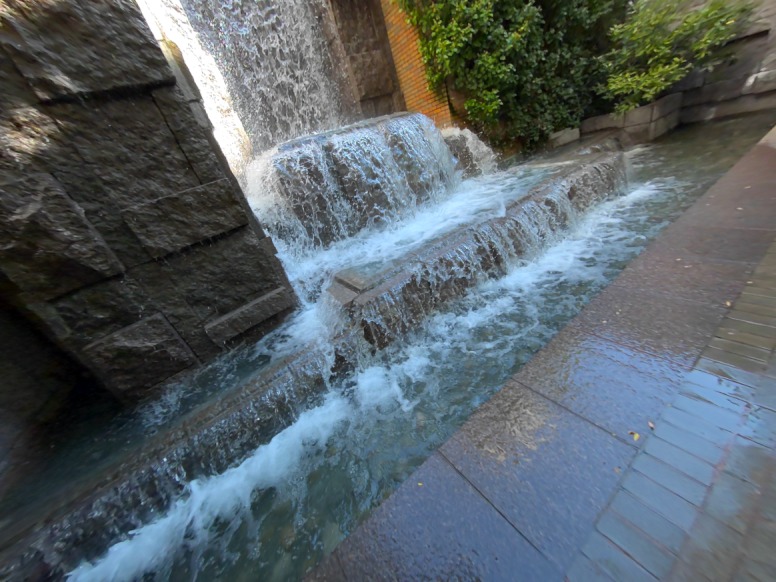}
    \caption{\textbf{Greenacre Park}\\[1pt]\scriptsize 750 frames (660/90)\\ 7.0\,m baseline, 111$^\circ$}
    \label{fig:scene-acre}
  \end{subfigure}
  \hfill
  \begin{subfigure}[t]{0.485\columnwidth}
    \centering
    \includegraphics[width=\linewidth,trim={130 100 130 100},clip]{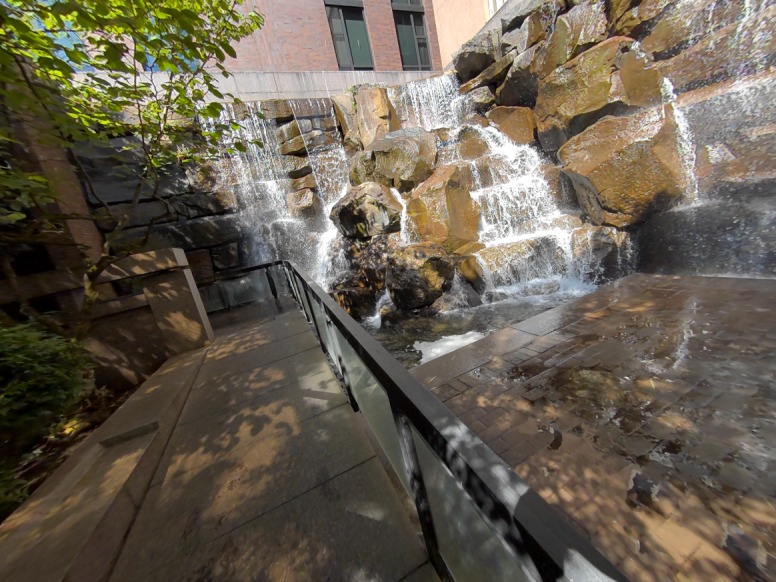}
    \caption{\textbf{Garden}\\[1pt]\scriptsize 750 frames (660/90)\\ 6.8\,m baseline, 101$^\circ$}
    \label{fig:scene-garden}
  \end{subfigure}
  \\[4pt]
  \begin{subfigure}[t]{0.485\columnwidth}
    \centering
    \includegraphics[width=\linewidth,trim={130 100 130 100},clip]{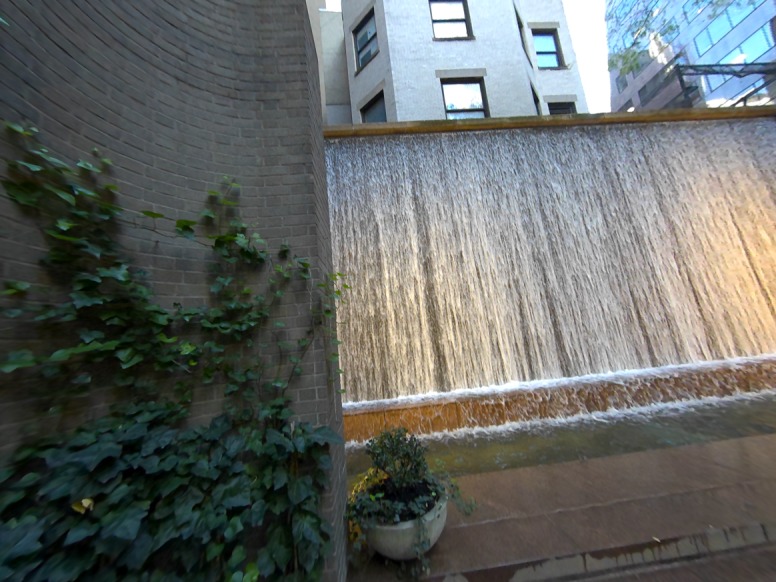}
    \caption{\textbf{Parley Falls}\\[1pt]\scriptsize 750 frames (660/90)\\ 8.4\,m baseline, 83$^\circ$}
    \label{fig:scene-parley}
  \end{subfigure}
  \hfill
  \begin{subfigure}[t]{0.485\columnwidth}
    \centering
    \includegraphics[width=\linewidth]{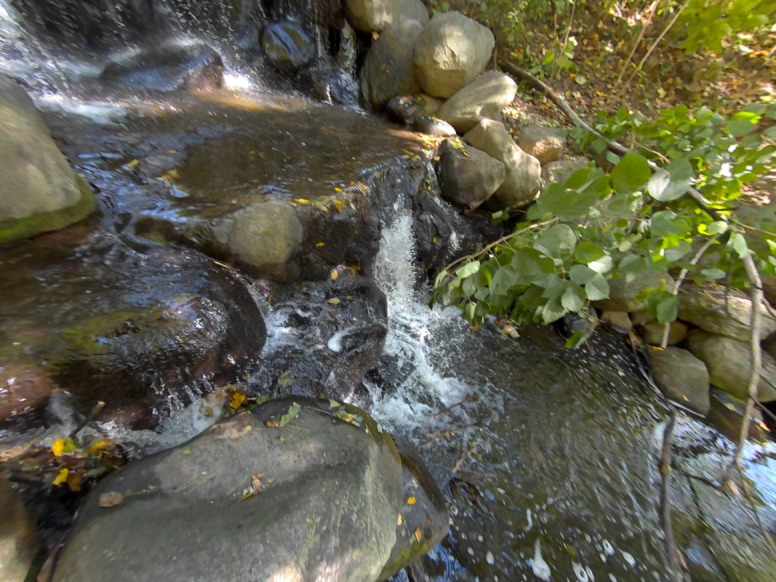}
    \caption{\textbf{Creek}\\[1pt]\scriptsize 750 frames (660/90)\\ 3.4\,m baseline, 93$^\circ$}
    \label{fig:scene-proc}
  \end{subfigure}
  \\[4pt]
  \begin{subfigure}[t]{0.485\columnwidth}
    \centering
    \includegraphics[width=\linewidth,trim={130 100 130 100},clip]{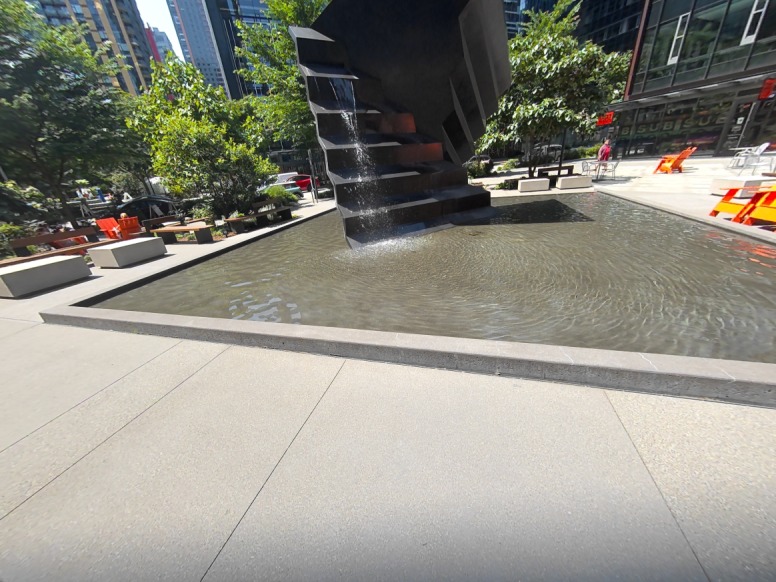}
    \caption{\textbf{Sculpture}\\[1pt]\scriptsize 750 frames (660/90)\\ 10.4\,m baseline, 105$^\circ$}
    \label{fig:scene-sculpture}
  \end{subfigure}
  \hfill
  \begin{subfigure}[t]{0.485\columnwidth}
    \centering
    \includegraphics[width=\linewidth]{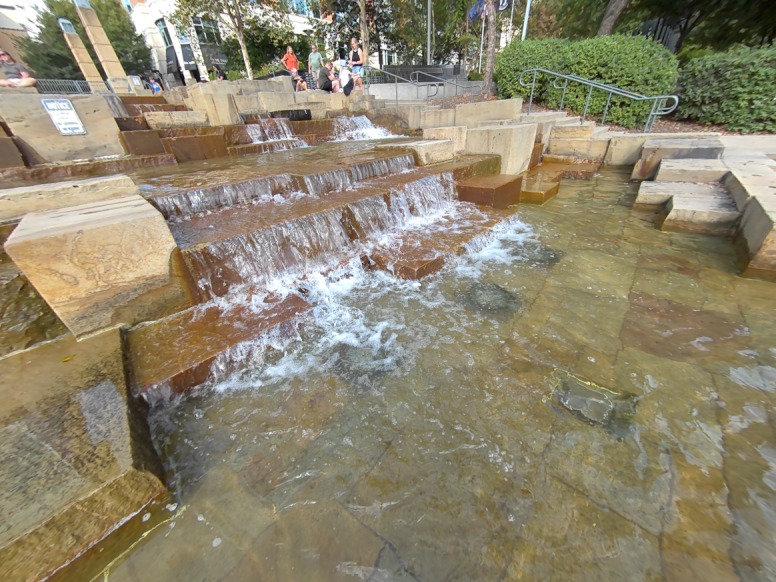}
    \caption{\textbf{Waterstep}\\[1pt]\scriptsize 800 frames (710/90)\\ 12.1\,m baseline, 101$^\circ$}
    \label{fig:scene-step}
  \end{subfigure}
  \\[4pt]
  \begin{subfigure}[t]{0.485\columnwidth}
    \centering
    \includegraphics[width=\linewidth]{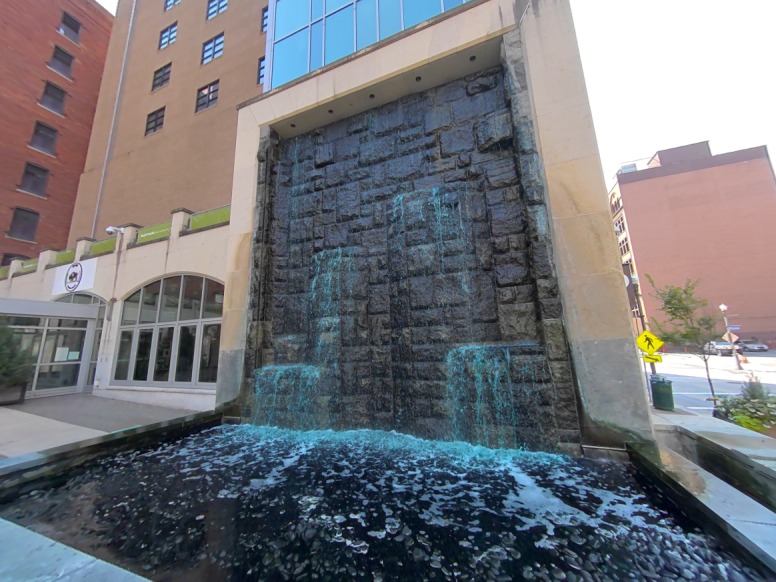}
    \caption{\textbf{Village Park}\\[1pt]\scriptsize 750 frames (660/90)\\ 9.4\,m baseline, 160$^\circ$}
    \label{fig:scene-vill}
  \end{subfigure}
  \caption{\textbf{The seven captured water scenes}, each shown at the middle frame of its sequence. Captions give the number of frames (train/test), the maximum camera baseline, and the maximum rotation between viewing directions. Every scene is a hand-held fisheye capture rectified to a pinhole camera; test frames are three held-out temporal segments per scene.}
  \label{fig:scenes}
\end{figure}

Here we clarify and provide additional details on the captured water scenes used in our experiments.
Figure~\ref{fig:scenes} shows one representative frame from each of the seven captured water scenes, and Tab.~\ref{tab:scenes} reports per-scene capture statistics.

\begin{table}[t]
  \centering
  \setlength{\tabcolsep}{3pt}
  \resizebox{\columnwidth}{!}{\small
  \begin{tabular}{lrrrrrrr}
    \toprule
    Scene & Frames & Train & Test & Baseline & Path & Rotation & Water \\
    \midrule
    Greenacre Park      & 750 & 660 & 90 & 7.0\,m  & 15\,m & 111$^\circ$ & 34\% \\
    Garden    & 750 & 660 & 90 & 6.8\,m  & 11\,m & 101$^\circ$ & 39\% \\
    Parley Falls    & 750 & 660 & 90 & 8.4\,m  & 17\,m & 83$^\circ$  & 39\% \\
    Creek      & 750 & 660 & 90 & 3.4\,m  & 11\,m & 93$^\circ$  & 38\% \\
    Sculpture & 750 & 660 & 90 & 10.4\,m & 14\,m & 105$^\circ$ & 44\% \\
    Waterstep      & 800 & 710 & 90 & 12.1\,m & 16\,m & 101$^\circ$ & 63\% \\
    Village Park      & 750 & 660 & 90 & 9.4\,m  & 17\,m & 160$^\circ$ & 34\% \\
    \bottomrule
  \end{tabular}}
  \caption{Per-scene capture statistics. \emph{Baseline} is the maximum distance between any two camera centres, \emph{Path} the total trajectory length, \emph{Rotation} the maximum angle between any two optical axes, and \emph{Water} the mean fraction of pixels inside the water mask.}
  \label{tab:scenes}
\end{table}

\section{Additional Ablations and Results}
\label{sec:supp_extra}

\subsection{Component Ablation Visualization}
\label{sec:supp_abl_tracks}

\begin{figure}[t]
  \centering
  {\setlength{\tabcolsep}{0pt}
   \newcommand{\ablhdr}[1]{{\fontfamily{ptm}\selectfont\scriptsize #1}}
   \begin{tabular}{@{}*{5}{>{\centering\arraybackslash}p{0.2\columnwidth}}@{}}
     \ablhdr{GT} & \ablhdr{Full Model} & \ablhdr{No initialization} & \ablhdr{\shortstack[c]{No non-periodic\\[-1pt]residual field}} & \ablhdr{\shortstack[c]{No Eulerian\\[-1pt]motion field}} \\
   \end{tabular}}\\[1pt]
  \includegraphics[width=\columnwidth]{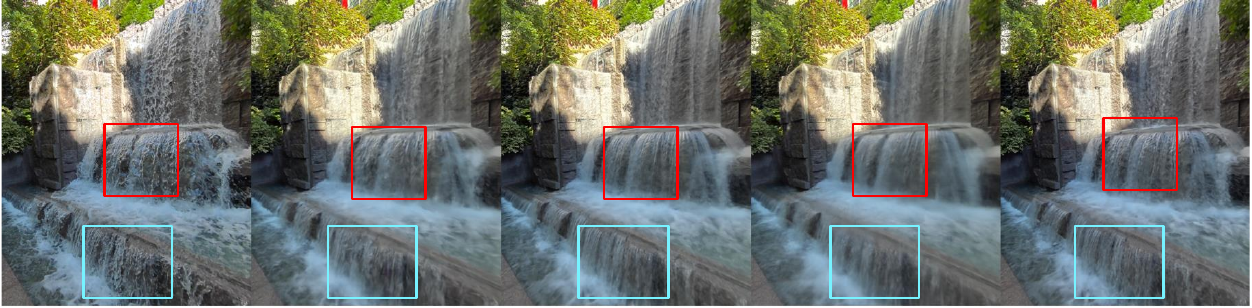}
  \vspace{-18pt}
  \caption{\textbf{Component ablations.} Our full model reconstructs the best dynamism of water with all the proposed components. Without initializing the motion field or learning a position-based deformation field from scratch converges to very small motion (water particles become blurry strands), while no opacity and color deformation prevents our looping representation from fitting to the non-looping input video, producing blurry results.}
  \label{fig:abl_tracks}
\end{figure}

Figure~\ref{fig:abl_tracks} shows a per-component reconstruction comparison on the \emph{acre} scene, complementing the track visualization in the main paper (Fig.~\ref{fig:ablation}). The full model reconstructs the best dynamism of water with all the proposed components. Without initializing the motion field or learning a position-based deformation field from scratch converges to very small motion (water particles become blurry strands), while no non-periodic residual prevents our looping representation from fitting to the non-looping input video, producing blurry results.

\section{User Study Details}
\label{sec:supp_user}

\begin{figure*}[t]
  \centering
  \begin{subfigure}[t]{\textwidth}
    \centering
    \includegraphics[width=\textwidth]{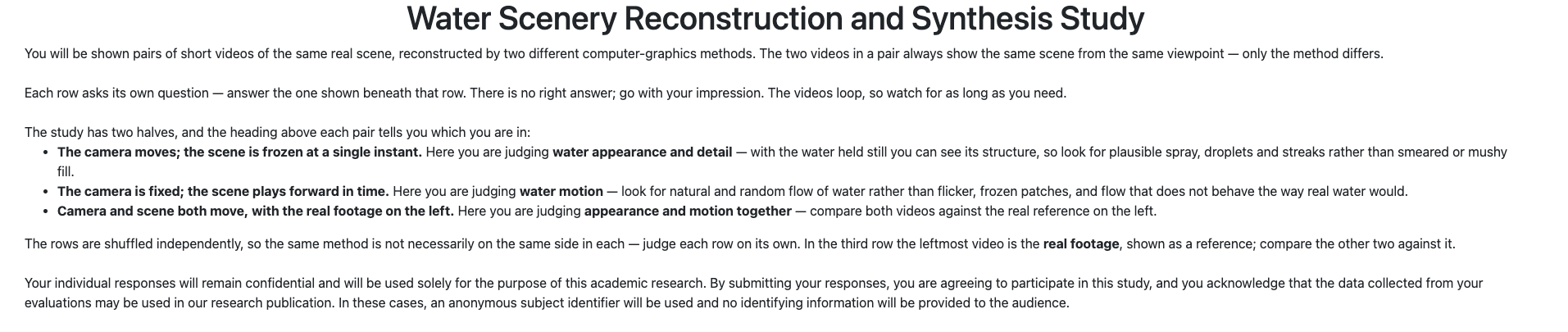}
    \caption{\textbf{Instructions.} Text shown to each of the $25$ participants at the start of the study.}
    \label{fig:supp_instructions}
  \end{subfigure}\\[6pt]
  \begin{subfigure}[t]{\textwidth}
    \centering
    \includegraphics[width=\textwidth]{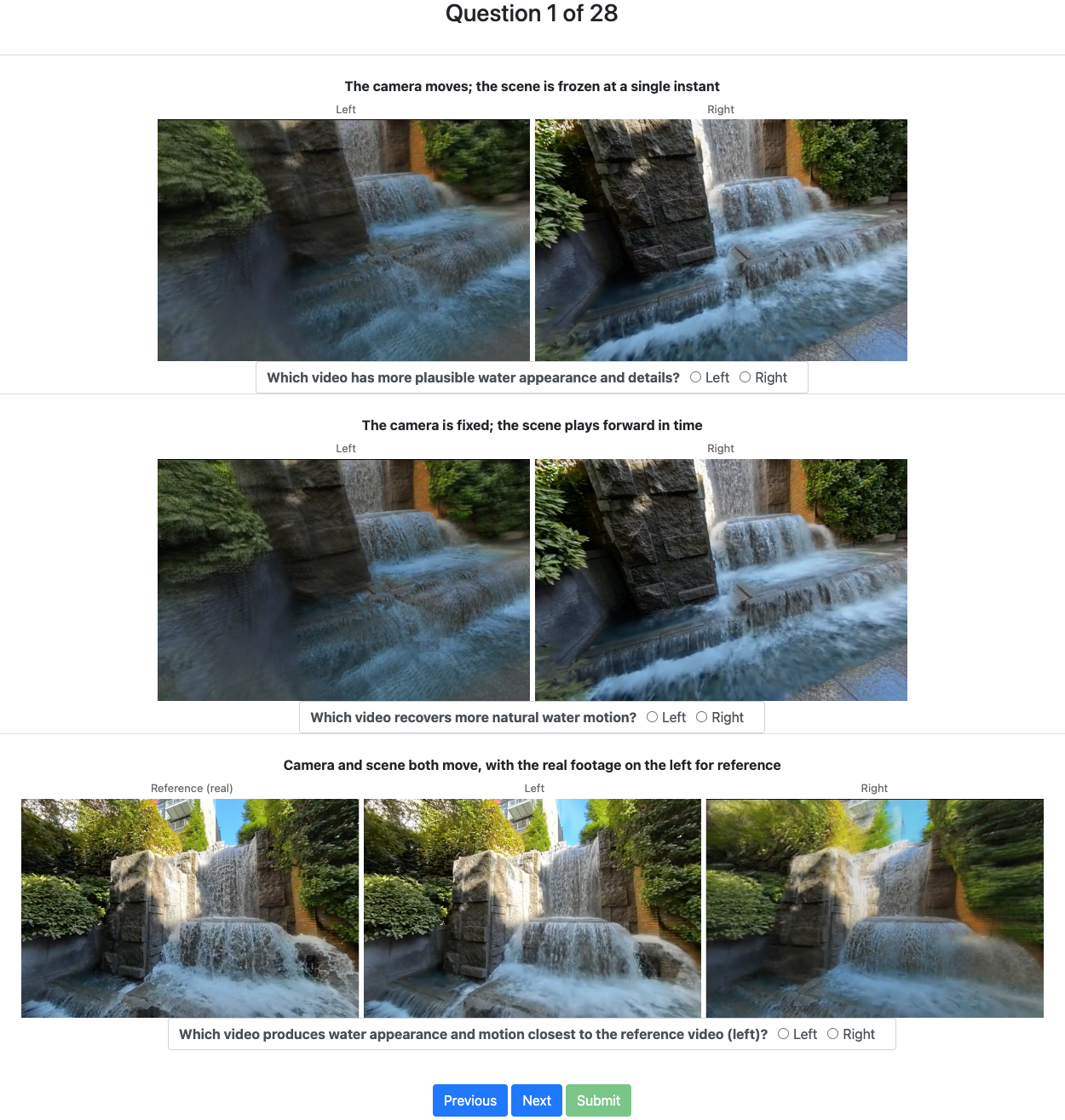}
    \caption{\textbf{Questions.} The three forced-choice questions asked on every comparison page, one per setting (\emph{Appearance}, \emph{Motion}, \emph{Reference}).}
    \label{fig:supp_questions}
  \end{subfigure}
  \caption{\textbf{User-study interface.} (a) instructions shown at the start of the study and (b) forced-choice questions asked on every comparison page.}
  \label{fig:supp_userstudy}
\end{figure*}

Figure~\ref{fig:supp_userstudy} shows (a) the instructions presented to each participant at the start of the study and (b) the three forced-choice questions asked on each comparison page (one per setting: appearance (fixed-time animation), motion(fixed-view animation), reference (reconstruction of held-out test frames compared to ground truth)).

{
    \small
    \bibliographystyle{ieeenat_fullname}
    \bibliography{citation}
}

\end{document}